\documentclass[11pt]{article}

\usepackage[final]{acl}

\usepackage{times}
\usepackage{latexsym}

\usepackage[T1]{fontenc}

\usepackage[utf8]{inputenc}

\usepackage{microtype}

\usepackage{inconsolata}

\usepackage{graphicx}
\usepackage{xspace}
\usepackage{xcolor}
\definecolor{cvprblue}{rgb}{0.21,0.49,0.74}
\newcommand{\ie}{\textit{i.e.}\xspace}
\newcommand{\eg}{\textit{e.g.}\xspace}

\usepackage{booktabs}
\usepackage{arydshln}
\usepackage[accsupp]{axessibility}
\usepackage{cuted}
\usepackage{wrapfig}
\usepackage{algorithm}
\usepackage{algpseudocode}

\usepackage{multirow}
\usepackage{amssymb} 
\usepackage{tabularx} 
\usepackage{booktabs} 
\usepackage{array}
\usepackage{subcaption}
\usepackage{makecell}
\usepackage{caption}
\usepackage{tikz} 
\newcolumntype{M}{>{\centering\arraybackslash}m{\dimexpr\hsize-2\tabcolsep}}
\usepackage[export]{adjustbox}
\usepackage{enumitem}

\usepackage{stfloats}

\makeatletter
\newcommand{\tightalgorithmbottom}[1][0.3cm]{%
  \global\def\@fs@post{\kern2pt\hrule\relax\vskip -#1}%
}
\makeatother

\hypersetup{
  breaklinks=true,
  colorlinks=true,
  allcolors=cvprblue,
  pdftitle={Cost-efficient Active Learning for Referring Image Segmentation and Grounding},
  pdfauthor={Junbeom Hong, Seonghoon Yu, Hyung Rok Jung, Sundong Kim, Jeany Son}
}

\usepackage{float}
\floatstyle{ruled}
\restylefloat{algorithm}

\title{Cost-efficient Active Learning for Referring Image Segmentation and Grounding}

\author{
  \textbf{Junbeom Hong}\thanks{Equal contribution.}\textsuperscript{,1}\qquad
  \textbf{Seonghoon Yu}\footnotemark[1]\textsuperscript{,2}\qquad
  \textbf{Hyung Rok Jung}\textsuperscript{3}
  \\
  \textbf{Sundong Kim}\textsuperscript{\textbf{3}}\qquad
  \textbf{Jeany Son}\textsuperscript{\textbf{4}}
\\[6pt] 
  \textsuperscript{1}Meissa\qquad
  \textsuperscript{2}KAIST\qquad
  \textsuperscript{3}GIST\qquad
  \textsuperscript{4}POSTECH
\\
  \small{
    jbhong@meissa.ai \quad seonghoon.yu@kaist.ac.kr \quad jhrock2001@gm.gist.ac.kr \quad sundong@gist.ac.kr \quad jeany@postech.ac.kr
  }
}

\begin{document}
\maketitle

\begin{abstract}

Collecting natural-language referring expressions along with region annotations, such as masks or boxes, is a major bottleneck in visual grounding (VG), as annotators must write descriptions that distinguish target regions from visually similar ones.
 We tackle this by formulating active learning (AL) for VG under the realistic setting where only raw images are available without accompanying text.
Since ground-truth text is unavailable, sample selection must estimate which images contain ambiguous regions that would require discriminative referring expressions.
To address this, we generate auxiliary region-text pairs using foundation models, and introduce Referred Region Ambiguity, a new acquisition function that measures whether the model's confidence collapses onto a single region or disperses across multiple candidates.
It allows our method to prioritize images with strong cross-region competition, which are more informative due to their visual ambiguity.
We also design a referring-expression annotation interface that helps annotators quickly focus on writing discriminative language with a few clicks.
Experiments on RIS and REC benchmarks show that our AL framework consistently outperforms several AL baselines, while a user study shows up to 1.6$\times$ faster description labeling of ours.
Code is available at \url{https://github.com/junbum766/ALRIS}.



\end{abstract}
\section{Introduction}
\label{sec:intro}
\vspace{-0.1cm}

Deep learning models often require large amounts of high-quality human annotations, making data collection a major bottleneck. 
Active learning (AL) addresses this issue by selecting the most \textit{informative} unlabeled samples for annotation, reducing labeling costs while maintaining performance. 
It has been widely applied to computer vision tasks such as image classification~\cite{bald, coreset}, object detection~\cite{ppal, boxlevel}, and semantic segmentation~\cite{al_seg_sup, al_seg_mul}.

\begin{figure}[t]
    \centering
    \includegraphics[width=1.0\linewidth]{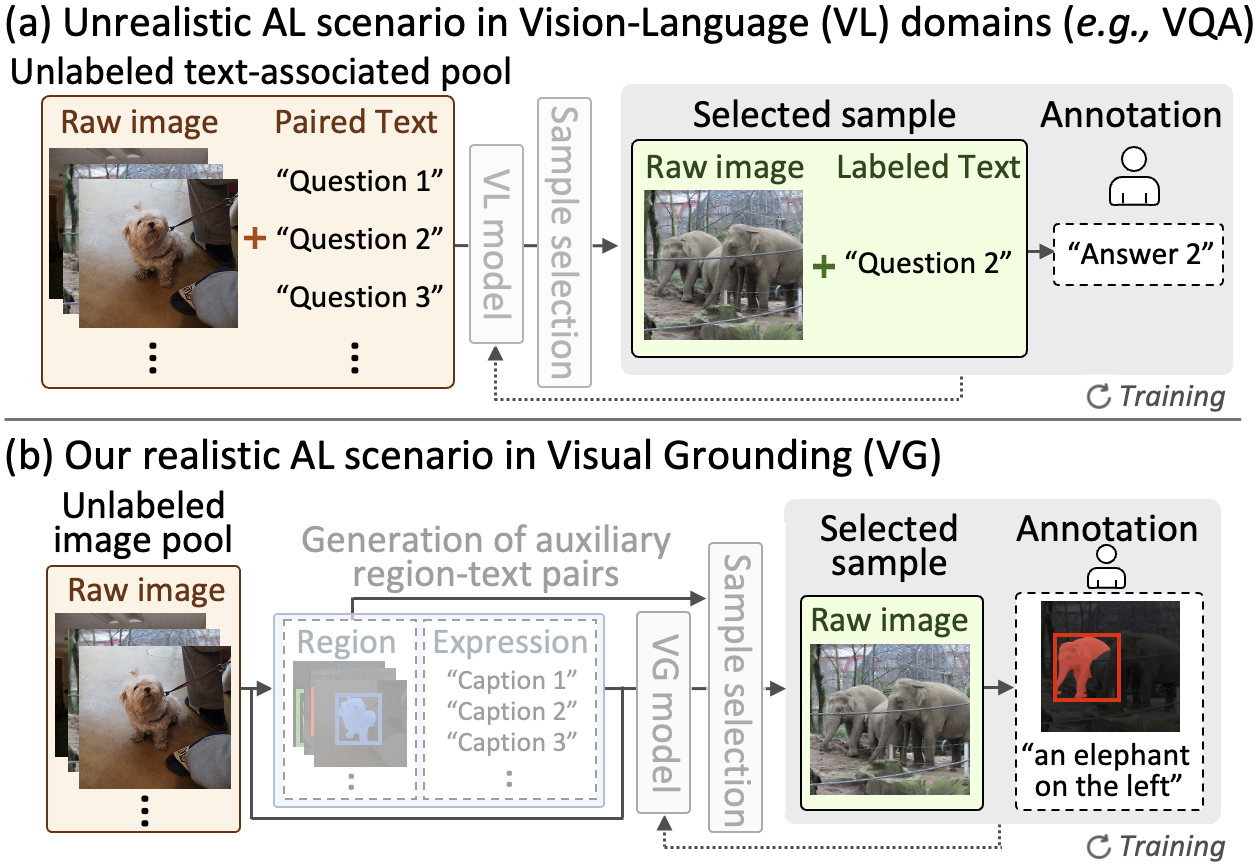}
    \vspace{-0.6cm}
    \caption{Illustration of AL settings in vision-language domains: (a) existing \textit{unrealistic} scenario~\cite{vqa_active, active_learning_vqa} in VQA assumes that costly textual questions related to each image are already available in the unlabeled pool, whereas (b) our \textit{realistic} setting in VG considers a practical scenario where only raw images are provided without textual descriptions.} 
    \vspace{-0.6cm}
    \label{fig:intro} 
\end{figure}

\begin{figure*}[t]
    \centering
    \begingroup
    \newcommand{\FixedImgB}[2]{%
      \begin{minipage}[c][#1][c]{\linewidth}
        \centering
        \includegraphics[width=\linewidth,height=#1,keepaspectratio]{#2}
      \end{minipage}%
    }
    \setcounter{subfigure}{0}
    \hfill
    \hspace{-0.2cm}
    \begin{subfigure}[t]{0.45\linewidth}
        \centering
        \setlength{\abovecaptionskip}{-2pt}
        \FixedImgB{0.12\textheight}{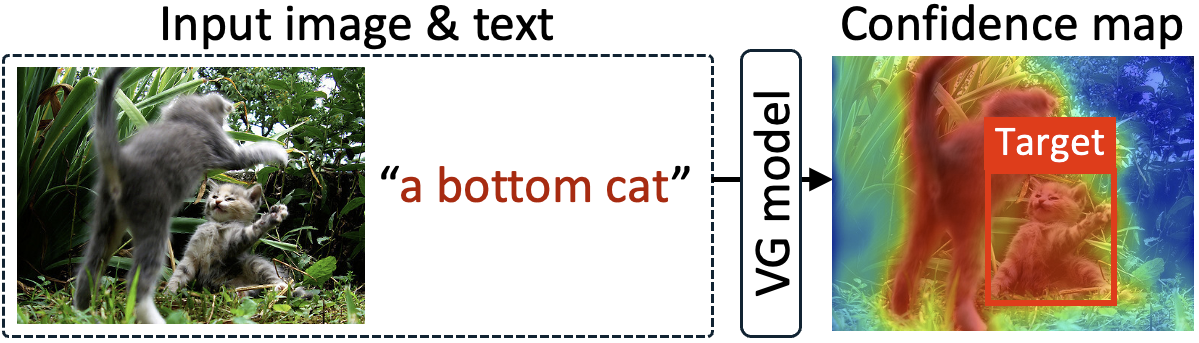}
        \vspace{-0.2cm}
        \caption{Informative example}
        \label{fig:intro2_a}
    \end{subfigure}
    \hfill
    \begin{subfigure}[t]{0.45\linewidth}
        \centering
        \setlength{\abovecaptionskip}{-2pt}
        \FixedImgB{0.12\textheight}{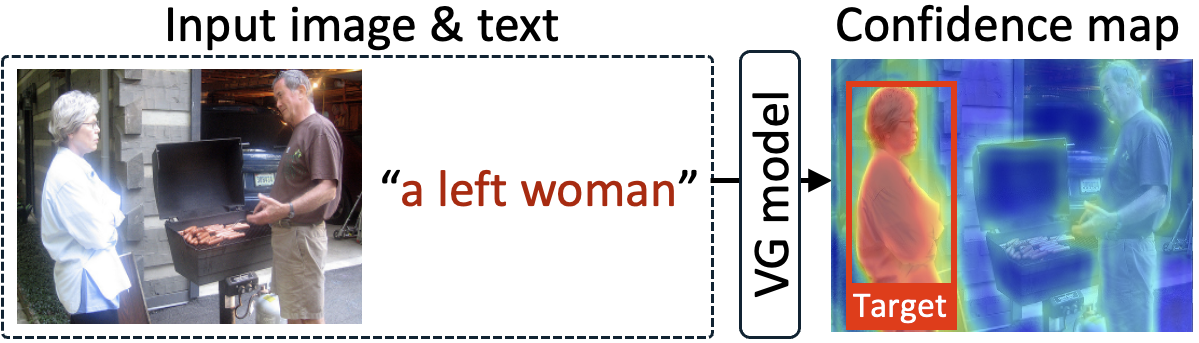}
        \vspace{-0.2cm}
        \caption{Non-informative example}
        \label{fig:intro2_b}
    \end{subfigure}
    \hspace{-0.2cm}
    \hfill
    \vspace{-0.25cm}
    \caption{Illustration of informativeness in VG: (a) the model assigns high confidence to multiple regions, indicating an informative sample, while (b) a referred one is solely activated by the model, showing low informativeness.}
    \vspace{-0.55cm}
    \label{fig:intro2}
    \endgroup
\end{figure*}


Despite its success in vision-only tasks, active learning (AL) remains underexplored for text-conditioned vision-language (VL) tasks such as visual question answering (VQA) and visual grounding (VG). 
Existing AL studies mainly focus on VQA and assume that image-text pairs are already available in the unlabeled pool~\cite{vqa,active_learning_vqa,vqa_active}, requiring annotators only to provide answer labels (Fig.~\ref{fig:intro}\textcolor{cvprblue}{a}).
This assumption is often unrealistic: real-world unlabeled data typically consists of raw images, where both linguistic labels, such as questions in VQA or referring expressions in VG, and their corresponding labels (answers or referred regions) must be annotated.
Moreover, AL for VG tasks such as referring image segmentation (RIS) and referring expression comprehension (REC) has not been explored despite their high annotation cost.


Another challenge in VG is that informativeness is characterized differently than in standard vision-only tasks.
In VG, the model must pinpoint one referred region among many visually or semantically similar candidates.
Therefore, when the model assigns high confidence to multiple regions simultaneously, such samples are, in fact, highly informative: they reveal exactly where the model lacks fine-grained discriminative ability (Fig.~\ref{fig:intro2}).
However, existing acquisition functions designed for segmentation or object detection fail to capture this property, since they primarily select samples with moderate overall confidence (\eg, pixel entropy~\cite{pixel-entropy}, box-level confidence~\cite{active_detection_entropy}).
As a result, cases where the model strongly activates multiple regions are mistakenly treated as low-informative, although they correspond to informative VG samples.

In this paper, we introduce a new active learning task for visual grounding that selects and annotates samples directly from raw images, reducing manual effort by labeling only a small number of highly informative samples (Fig.~\ref{fig:intro}\textcolor{cvprblue}{b}).
First, to address the lack of textual descriptions in the unlabeled image pool, we generate auxiliary region-text pairs, providing the required multi-modal input for a VG model and sample selection. 
Second, to capture VG-specific informativeness, we introduce a novel acquisition function termed \textit{Referred Region Ambiguity}. 
This metric measures how the model's confidence is distributed across multiple candidate regions instead of being concentrated in one referred region.
Its high ambiguity indicates that multiple regions are similarly plausible under the same expression, exposing genuine ambiguity that is worth labeling. 
By prioritizing such ambiguous samples, our framework enables the VG model to rapidly improve its fine-grained discriminative capability with minimal annotation costs.


Furthermore, we propose an efficient referring-expression annotation interface along with mask annotations for VG. 
Our interface first lets annotators specify the target region, then assists expression writing with region-aware top-$k$ word suggestions. 
Rather than typing full sentences from scratch, annotators construct distinct referring expressions by sequentially selecting suggested top-$k$ words relevant to the target region with a few clicks, achieving a significant reduction in the labeling cost.



Under the realistic AL scenarios in VG, our active learning framework achieves state-of-the-art performance over several AL baselines on both RIS and REC benchmarks.
Notably, the proposed labeling tool achieves up to 1.6$\times$ faster labeling speeds for referring expression annotations, compared to the na\"ive labeling in a user study.

Our contributions can be summarized as follows:

\begin{itemize}[leftmargin=10pt, itemsep=0pt, topsep=0pt, parsep=2pt, partopsep=2pt]
    \item We introduce a new task setting for active learning in visual grounding, where sample selection must be performed from raw images without pre-existing image-text pairs, reflecting a realistic data collection scenario.

    \item We propose a novel acquisition, \textit{Referred Region Ambiguity}, built on auxiliary region-text pairs, which selects ambiguous samples where the model assigns high confidence to multiple regions rather than a referred one, enabling faster learning of fine-grained discrimination.

    \item We present an efficient referring-expression labeling tool with region annotations that reduces annotation cost by reformulating expression construction as a word-by-word click process.

    \item Our approach surpasses several AL baselines on RIS and REC benchmarks while concurrently delivering superior labeling efficiency in both speed and annotation cost.
\end{itemize}

\begin{figure*}[t]
    \centering
    \includegraphics[width=1.0\linewidth]{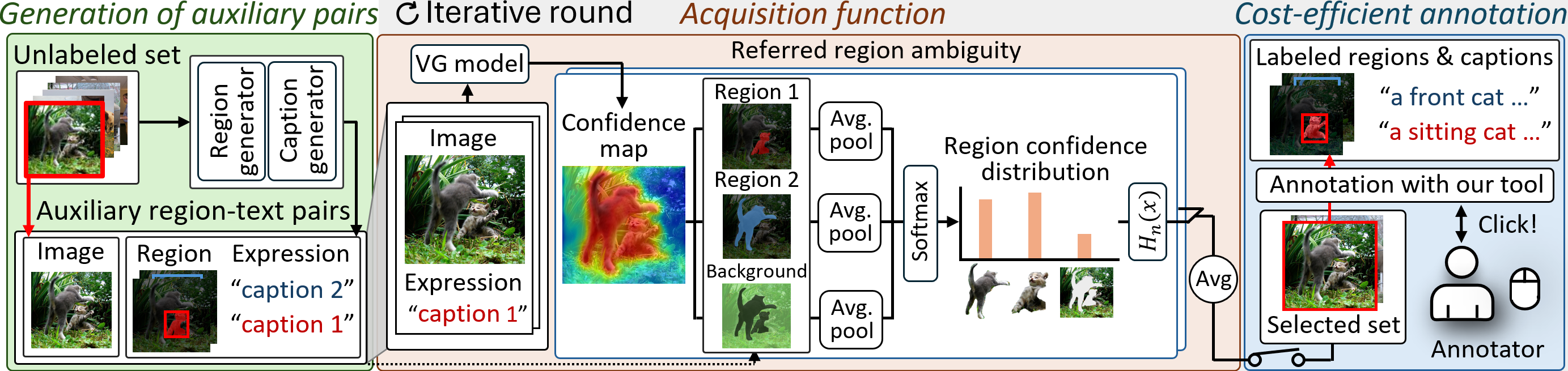}
    \vspace{-0.6cm}
    \caption{The proposed AL framework for visual grounding. We first generate auxiliary region-text pairs for all raw images in the entire unlabeled pool (Sec.~\ref{subsec:pseudo_data}).
    Following this, iterative rounds of sample selection and labeling begin.
    In each round, our referred region ambiguity (Sec.~\ref{subsec:acquisition}) scores the informativeness of all raw images by utilizing their auxiliary region-expression pairs. The top-scoring images are then sequentially annotated via our cost-efficient interface (Sec.~\ref{subsec:annotation_tool}) until the round budget is exhausted.
    }
    \vspace{-0.55cm}
    \label{fig:overall_framework} 
\end{figure*}

\vspace{-0.15cm}
\section{Related Work}
\label{sec:related}
\vspace{-0.15cm}

\paragraph{Active Learning.}
Active learning (AL) reduces annotation costs by selecting informative samples from unlabeled data pools.
Prior work has explored uncertainty~\cite{LeastConfidence, bald, lpl, yeho}, diversity~\cite{badge}, and hybrid~\cite{dbal, waal} criteria in image classification, and then extended to detection~\cite{active_detection_entropy, boxlevel}, segmentation~\cite{al_seg_foundation, al_seg_adaptive_superpixel, al_seg_cor}, and VQA~\cite{vqa, active_learning_vqa, vqa_active}.
However, VL AL typically assumes pre-existing language inputs.
Although ACTRESS~\cite{actress} actively selects pseudo labels for semi-supervised VG from image-text pairs, it does not address human annotation cost incurred in text modality.
In contrast, we study realistic active VG data collection from raw images, where both target regions and referring expressions are newly annotated.

\vspace{-0.15cm}

\paragraph{Cost-efficient Annotation Tool.}
Reducing per-sample labeling cost is crucial for AL efficiency.
Prior works have reduced visual annotation costs for segmentation using superpixels~\cite{al_seg_metabox+, al_seg_mul} or foundation models~\cite{al_seg_cor, al_seg_foundation}, but mainly target mask annotation.
In contrast, referring expression annotation remains largely unexplored despite a bottleneck in VG.
To this end, we introduce a first interface for referring expressions, together with referred region annotations.




\vspace{-0.15cm}

\paragraph{Visual Grounding.}
Visual grounding (VG) localizes regions from language, producing masks for RIS~\cite{risclip, sharedris} or boxes for REC~\cite{related_rec_transvg++, related_rec_grounding_dino}.
Since VG annotation requires both regions and detailed expressions, prior work has reduced supervision via weakly-\cite{shatter_gather, weak-rec}, semi-supervised~\cite{semi-rec, semi-ris}, zero-shot~\cite{zero-shot-ris}, and pseudo-labeling~\cite{pseudo-q, pseudoris} approaches.
However, their gap to fully supervised methods~\cite{latentris, oneref} highlights the value of human labels.
In this work, we explore active learning for VG to minimize annotation cost while preserving human-label quality.

\vspace{-0.1cm}
\section{Active Learning for Visual Grounding}
\vspace{-0.15cm}

We introduce an active learning framework for visual grounding that reduces manual labeling costs by selecting a small subset of informative raw images. 
Our goal is to identify challenging samples where the model struggles to ground an expression to the intended region among similar candidates. 
Such ambiguity reveals the fine-grained discrimination required for VG. 
We describe the task setup, auxiliary region-text pair construction, and our acquisition function in this section.

{%
\begin{algorithm*}[t]\tightalgorithmbottom[0.5cm] 
    \caption{Proposed Active Learning Framework for VG}
    \label{algorithm:1}
    \begin{algorithmic}[1]
    \vspace{-0.1cm}
        \Require Unlabeled image set $\mathcal{I}$, the number of AL rounds $R$, budget per round $B_r$, an initial model $\theta_0$
            \State Build a set of auxiliary pairs $\mathcal{D}_{\text{aux}}$~on $\mathcal{I}$ \Comment{Sec.~\ref{subsec:pseudo_data}}
            \For{$r = 1,2,\dots,R$}
                \State Score informativeness of all raw images $\forall I \in \mathcal{I}$ via $\theta_{r-1}$ and $\mathcal{D}_{\text{aux}}$ \Comment{Sec.~\ref{subsec:acquisition}}
                \State Annotate top-scored images with our annotation interface \Comment{Sec.~\ref{subsec:annotation_tool}}
                \State Build a set of labeled images $\mathcal{D}_{r}$ within budget $B_r$
                \State Get the model $\theta_{r}$ trained on labeled dataset $\bigcup_{i=1}^r\mathcal{D}_i$
            \EndFor
            \State \textbf{Return} The final model $\theta_R$
    \end{algorithmic}
\end{algorithm*}
}


\vspace{-0.15cm}
\subsection{Overall Framework}
\vspace{-0.1cm}
We start by formalizing the active learning task in the context of visual grounding under a realistic setting, where only raw images are available in an unlabeled pool.
Formally, let $\mathcal{I}=\{I_1, \ldots, I_{|\mathcal{I}|}\}$ denote an unlabeled image set.
The goal is to iteratively select a subset of informative images $\{I_n\}_{n=1}^{N_r}\subset\mathcal{I}$ for human labeling at each round $r$.
To enable VG sample selection on raw images, our framework begins with \textit{generation of auxiliary region-text pairs} (Sec.~\ref{subsec:pseudo_data}), where it produces auxiliary pairs for each text-less image to infer the informativeness.

In every round $r$, our \textit{acquisition function} (Sec.~\ref{subsec:acquisition}) assesses informativeness scores of every unlabeled image using its corresponding auxiliary samples.
Based on these scores, the top informative raw images $\{I_n\}_{n=1}^{N_r}\subset\mathcal{I}$ are sequentially annotated by humans via our \textit{cost-efficient interface} (Sec.~\ref{subsec:annotation_tool}) until the round's budget $B_r$ is exhausted.
These newly labeled samples $\mathcal{D}_{r}$ are added to the manually labeled set $\bigcup_{i=1}^{r-1}\mathcal{D}_i$, and then the VG model $\theta_{r-1}$ is retrained on these updated labeled data $\bigcup_{i=1}^{r}\mathcal{D}_i$ to begin the next AL round.
The overall procedure is summarized in Algorithm~\ref{algorithm:1} and illustrated in Fig.~\ref{fig:overall_framework}.
In the following sections, we further elaborate on each step in more detail.

\vspace{-0.15cm}
\subsection{\texorpdfstring{\textls[-12]{Generation of Auxiliary Region-Text Pairs}}{Generation of Auxiliary Region-Text Pairs}}
    \vspace{-0.1cm}
\label{subsec:pseudo_data}
A core challenge in applying AL to VG lies in estimating raw-image informativeness without textual descriptions.
We address this by generating a set of auxiliary region-text pairs for each unlabeled image, offering potential semantics to estimate its informativeness.
Specifically, for a given image $I$, we extract a set of $N$ potential regions $\{r^{\text{aux}}_{i}\}^{N}_{i=1}$ using Grounded-SAM~\cite{grounded_sam} guided by COCO 80 categories as semantic prompts,
where $N$ depends on the number of detected regions.

Next, we generate corresponding auxiliary captions $\{t^{\text{aux}}_{i}\}_{i=1}^{N}$ using ViP-LLaVA~\cite{vip_llava} for all regions, with each region highlighted by a red box to guide generation.
These region-caption pairs form $\mathcal{D}_{\text{aux}}$ over the unlabeled pool $\mathcal{I}$, which provide potential signals to score the informativeness of each raw image via the acquisition function in the next subsection.


    \vspace{-0.13cm}
\subsection{Acquisition Function}
    \vspace{-0.1cm}
\label{subsec:acquisition}
Our acquisition function, termed \textbf{Referred Region Ambiguity} (Fig.~\ref{fig:overall_framework}), is designed to capture the unique source of informativeness in VG: ambiguity among candidate referred regions.
While standard acquisition functions measure uncertainty at the pixel or box level~\cite{pixel-entropy,active_detection_entropy}, they overlook the cross-region competition induced by a referring expression.
In contrast, our method treats candidate regions as competing referents and computes a region-level confidence distribution for each auxiliary caption.
It assigns high scores when the model distributes confidence across multiple plausible regions, revealing its difficulty in identifying the intended referent, and low scores when confidence is concentrated on a single region.
This formulation directly targets the fine-grained discrimination required in VG and provides a more faithful informativeness signal than generic vision-only acquisition criteria that do not account for language-conditioned region ambiguity.


Given a raw image $I$ with its corresponding $N$ auxiliary pairs $\{(r^{\text{aux}}_i,t^{\text{aux}}_i)\}_{i=1}^N$, where $r_i^{\text{aux}}$ is an auxiliary region (\ie, a mask for RIS or a box for REC) and $t_i^{\text{aux}}$ is its corresponding caption, 
we quantify how exclusively the model attends to the intended region for each auxiliary caption.
For each caption $t_i^{\text{aux}}$, the model produces a confidence map (\ie, a final logit scoring map before thresholding for RIS or attention map from the final decoder layer for REC), $p^i= \theta(I, t_i^{\text{aux}}) \in \mathbb{R}^{h\times w}$. 
{
We then compute region-level confidence scores $s^i_j$ over $N+1$ regions ($N$ auxiliary regions plus a background region $r^{\text{aux}}_0$) by averaging the confidence values within each region, as follows:
}
{
\setlength{\abovedisplayskip}{5.5pt}
 \setlength{\belowdisplayskip}{5.5pt}
 \setlength{\abovedisplayshortskip}{5.5pt}
 \setlength{\belowdisplayshortskip}{5.5pt}
\begin{equation}
    s^i_j = \frac{1}{|r^{\text{aux}}_j|}\sum_{(x,y)\in r^{\text{aux}}_j}p^i_{(x,y)} \in \mathbb{R}, ~ \forall j\in\{0, \dots,N\}, \nonumber
\end{equation}
}where $p^i_{(x,y)}$ denotes the confidence score at the pixel coordinate $(x,y)$.
These scores are normalized with a softmax over all regions, producing a probability distribution $\mathbf{q}^i=\frac{\exp({s^i_j})}{\sum^{N}_{j=0}\exp({s^i_j})}\in\mathbb{R}^{N+1}$ for the $i$-th caption.
{We then compute the normalized entropy $H_n(\mathbf{q}^i)$ of $\mathbf{q}^i$, as follows:}
{
\setlength{\abovedisplayskip}{6pt}
 \setlength{\belowdisplayskip}{6pt}
 \setlength{\abovedisplayshortskip}{6pt}
 \setlength{\belowdisplayshortskip}{6pt}
\begin{align}
    H_{n}(\mathbf{q}^i) &=-\frac{\sum_{j=0}^Nq^i_j \log(q^i_j)}{\log(N+1)}\in \mathbb{R},
\end{align}
}{where $H_n(\mathbf{q}^i)$ is computed by taking the entropy of $\mathbf{q}^i$ and then dividing it by the maximum entropy value over $N+1$ regions, $\log(N+1)$.}
A higher value of $H_{n}(\mathbf{q}^i)$ indicates greater ambiguity, meaning the model assigns similar confidence to multiple candidate regions rather than focusing on a single dominant one.

\vspace{-0.19cm}
Finally, the informativeness of an image $I$ is then computed by averaging $H(\mathbf{q}^i)$ over all its auxiliary captions, as follows:
{
\setlength{\abovedisplayskip}{5.7pt}
 \setlength{\belowdisplayskip}{5.7pt}
 \setlength{\abovedisplayshortskip}{5.7pt}
 \setlength{\belowdisplayshortskip}{5.7pt}
\begin{align}
    E(I) = \frac{1}{N} \sum_{i=1}^NH_n(\mathbf{q}^i),
\end{align}
}which we define as \textit{Referred Region Ambiguity}.
Images with the highest $E(I)$ scores are then annotated with our cost-efficient labeling tool (Sec.~\ref{subsec:annotation_tool}) until the next AL round budget is exhausted.


    \vspace{-0.2cm}

\begin{figure*}[t]
    \centering
    \begin{subfigure}[b]{0.48\textwidth}
        \centering
        \includegraphics[width=\textwidth]{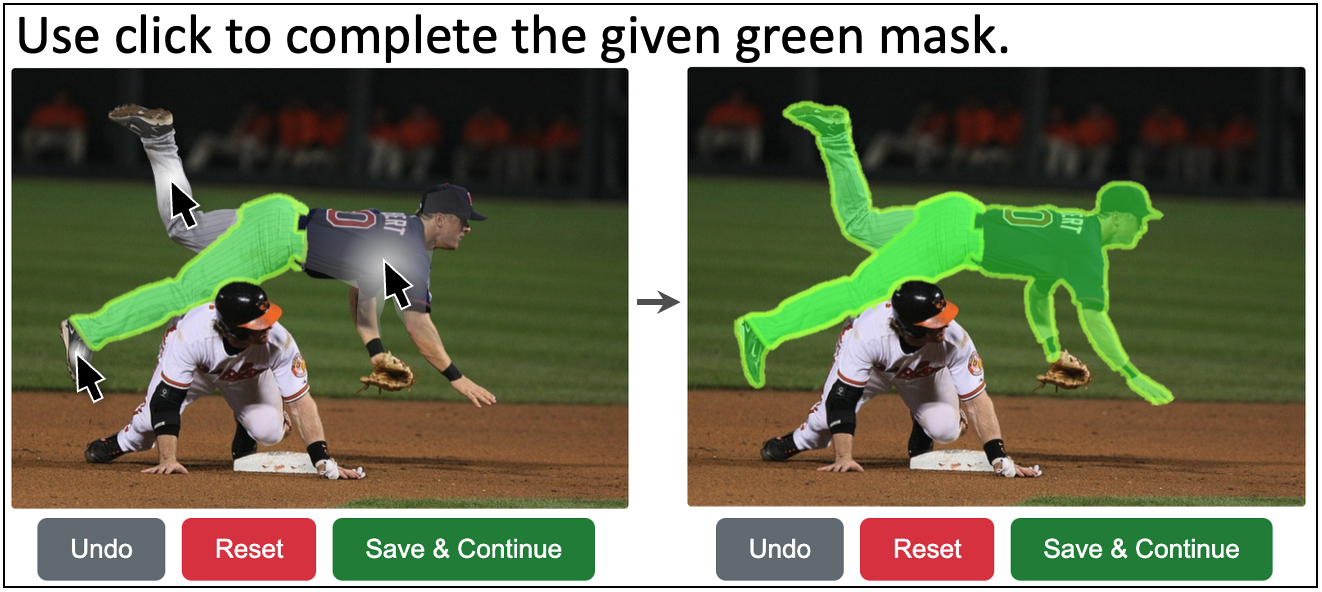}
        \caption{Target-Mask Specification}
        \label{fig:interface:a}
    \end{subfigure}
    \begin{subfigure}[b]{0.48\textwidth}
        \centering
        \includegraphics[width=\textwidth]{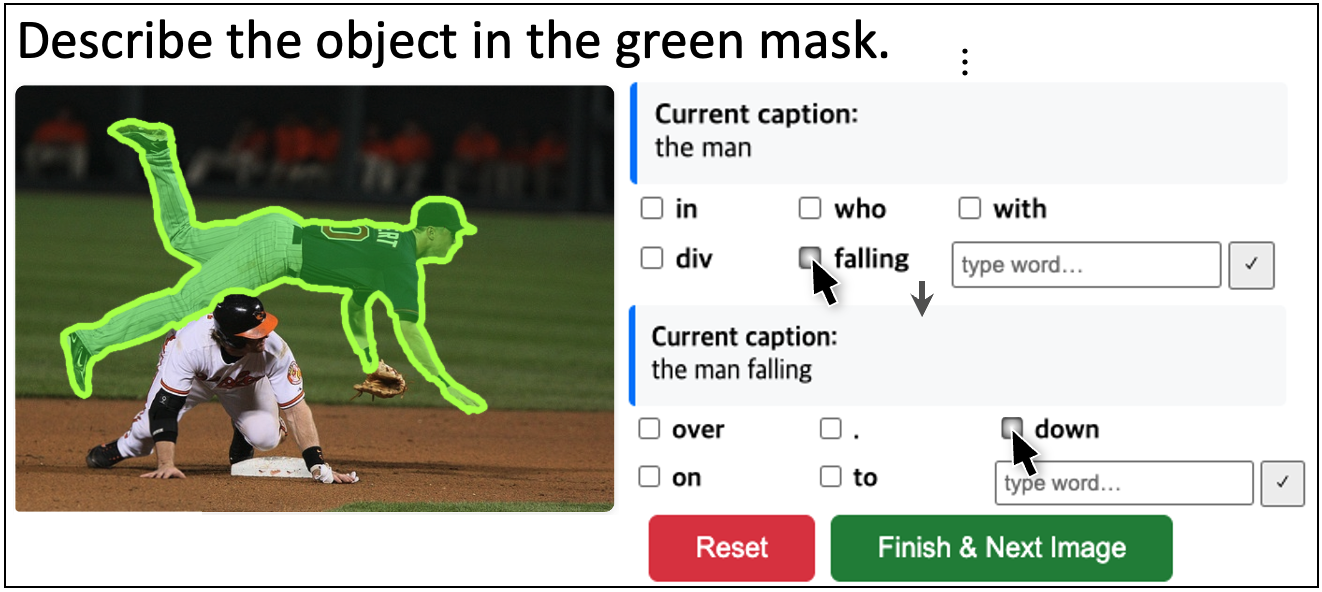}
        \caption{Referring-Expression Labeling Interface}
        \label{fig:interface:b}
    \end{subfigure}
    \vspace{-0.35cm}
    \caption{The proposed cost-efficient annotation interface. (a) target-mask specification, where the user refines a presented mask by clicking on a desired region in sequence, and 
    (b) text labeling interface, where the annotator composes a referring expression for the green mask by clicking on suggested word candidates sequentially, with the option to manually type a word if none of the suggested words are suitable. Demo videos are available in our \href{https://github.com/junbum766/ALRIS}{GitHub repository}.
    }
    \vspace{-0.55cm}
    \label{fig:interface}
\end{figure*}

\section{Cost-efficient Annotation Interface}
\label{subsec:annotation_tool}

\vspace{-0.2cm}

To mitigate annotation overhead in our AL pipeline, we further present a cost-efficient tool for labeling referring expressions. 
As shown in Fig.~\ref{fig:interface}, it enables click-driven target-region annotation and word-by-word expression construction via simple clicks, avoiding labor-intensive mask drawing and free-form text typing. 
We first detail our annotation workflow and then present its cost measurement.



\vspace{-0.25cm}

\paragraph{Target-Region Specification.}
We first specify the target region to be described by a referring expression.
For RIS, annotators start from an auxiliary mask from Sec.~\ref{subsec:pseudo_data} and refine it through simple clicks, each selecting and merging a pre-extracted SAM~\cite{sam} mask. 
For REC, we use two-click bounding-box specification.


\vspace{-0.25cm}

\paragraph{Referring-Expression Annotation.}
For each finalized region, annotators construct a referring expression word by word, as shown in Fig.~\ref{fig:interface:b}.
At each step, the interface presents $k$ suggested words, which annotators can select with a click or replace by choosing the \textit{typing} option.
This process repeats until the expression is complete.
The word candidates are generated by a region multimodal LLM (\ie, ViP-LLaVA~\cite{vip_llava}): at each auto-regressive decoding step, the model predicts the next-word distribution conditioned on the image, region, and previously selected words, and the top-$k$ words are shown as clickable candidates.


\vspace{-0.25cm}

\paragraph{Cost Measurement.}
We define each sample's labeling cost as the sum of its region cost (mask for RIS or box for REC) and text cost, computed as:
\\
1) \textbf{Target-region costs} ($C_{\text{region}}$):
for RIS, it is defined as the number of clicks required to finalize the mask labeling;
for REC, it is fixed to two clicks to specify the diagonal corners of a single box.
\\
2) \textbf{Referring-expression costs} ($C_{\text{exp}}$):
we define this cost as the sum of each word selection cost $C_{\text{w}}$ over an expression.
For each word selection, 
if the desired word appears in the top-$k$ suggestions with probability $p_k$, the user selects it from $k+1$ options, costing $\log_2(k+1)$ bits. 
Otherwise, the user selects the typing option with the same cost and additionally pays a typing cost $C_{\text{type}}$. $C_{\text{type}}$ is computed as the conditional information content of typed letters, estimated by a frequency-based letter-level 3-gram model.
Formally,
{
\setlength{\abovedisplayskip}{3pt}
 \setlength{\belowdisplayskip}{3pt}
 \setlength{\abovedisplayshortskip}{3pt}
 \setlength{\belowdisplayshortskip}{3pt}
\begin{align}
    \label{costmeasure}
    C_{\text{word}} = &\underbrace{\log_2(k+1)}_{\text{Click a word or typing option}}+ \underbrace{(1-p_k)C_{\text{type}}}_{\text{Type a word manually}}, \nonumber \\
    \text{whe}\text{re}~~&C_{\text{type}} = -\sum_{i=1}^n\log_2p_l(l_i|l_{i-1},l_{i-2}),
\end{align}
}and $p_l(l_i|l_{i-1},l_{i-2})$ is the conditional probability of letter $l_i$ given previous two letters.
Statistics for $p_k$ and $p_l(l_i|l_{i-1},l_{i-2})$ are in Appendix~\ref{supple:details_text_cost}.

\vspace{-0.25cm}

\begin{figure*}[t]
    \centering 

    \includegraphics[width=1\textwidth]{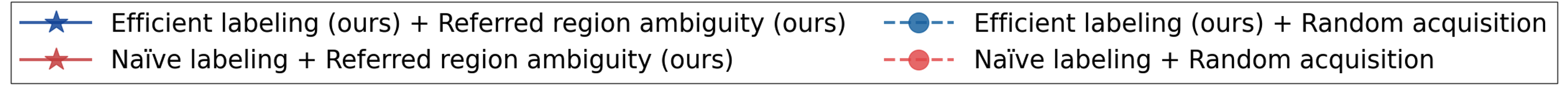}
    \begin{subfigure}{0.31\textwidth}
        \includegraphics[width=\linewidth]{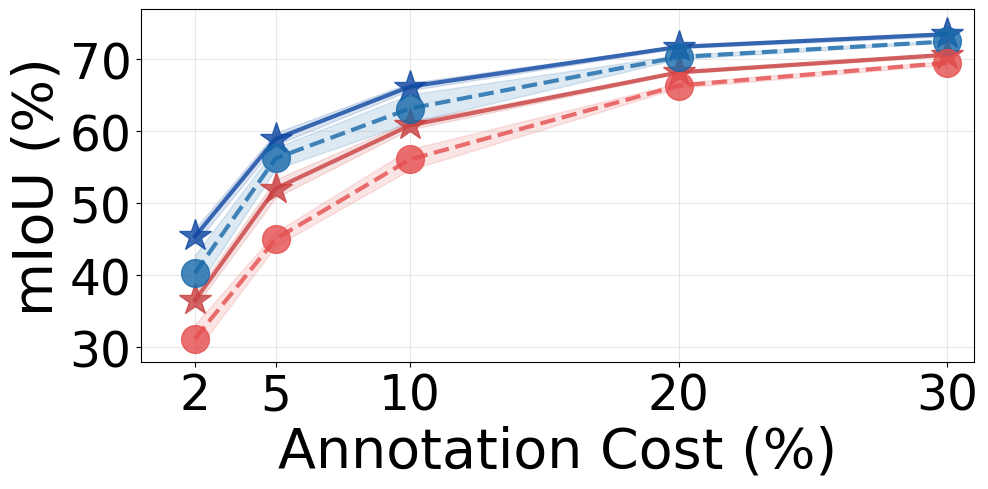}
        \vspace{-0.6cm}
        \caption{RefCOCO} 
        \label{fig:main_ris_a}
    \end{subfigure}
    \hfill 
    \begin{subfigure}{0.31\textwidth}
        \includegraphics[width=\linewidth]{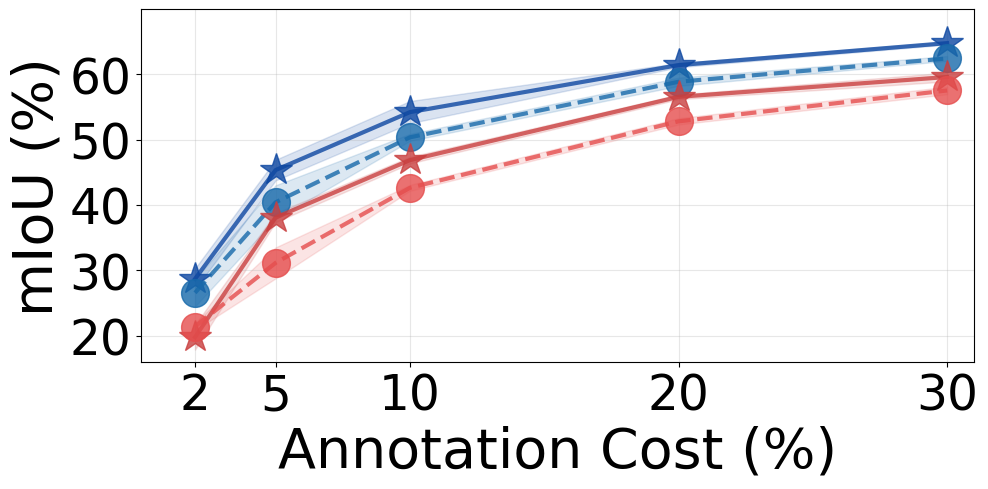}
        \vspace{-0.6cm}
        \caption{RefCOCO+} 
        \label{fig:main_ris_b}
    \end{subfigure}
    \hfill 
    \begin{subfigure}{0.31\textwidth}
        \includegraphics[width=\linewidth]{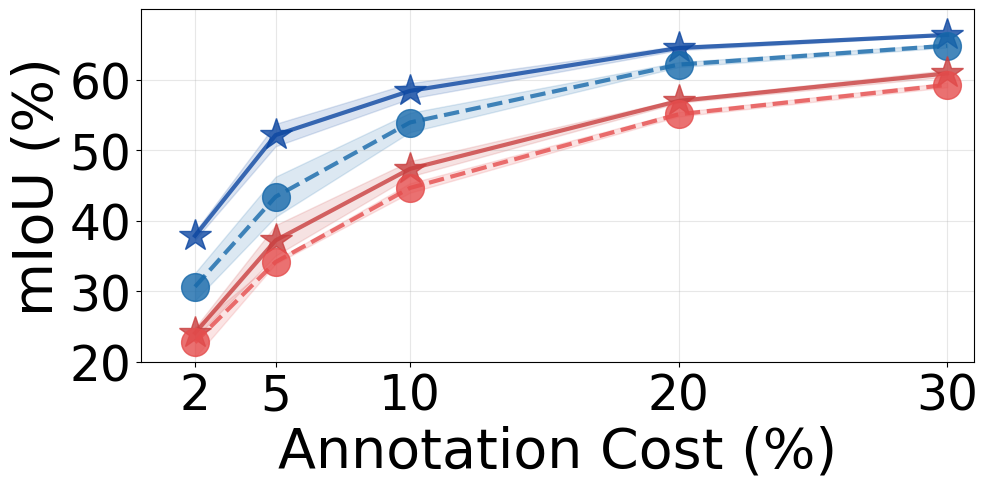}
        \vspace{-0.6cm}
        \caption{RefCOCOg} 
        \label{fig:main_ris_c}
    \end{subfigure}
    \vspace{-0.3cm}
    \caption{Comparison with the na\"ive labeling and random acquisition on a \textbf{RIS} task.}
    \vspace{-0.2cm}
    \label{fig:main_ris_figure}
\end{figure*}

\begin{figure*}[t]
    \centering 

    \begin{subfigure}{0.31\textwidth}
        \includegraphics[width=\linewidth]{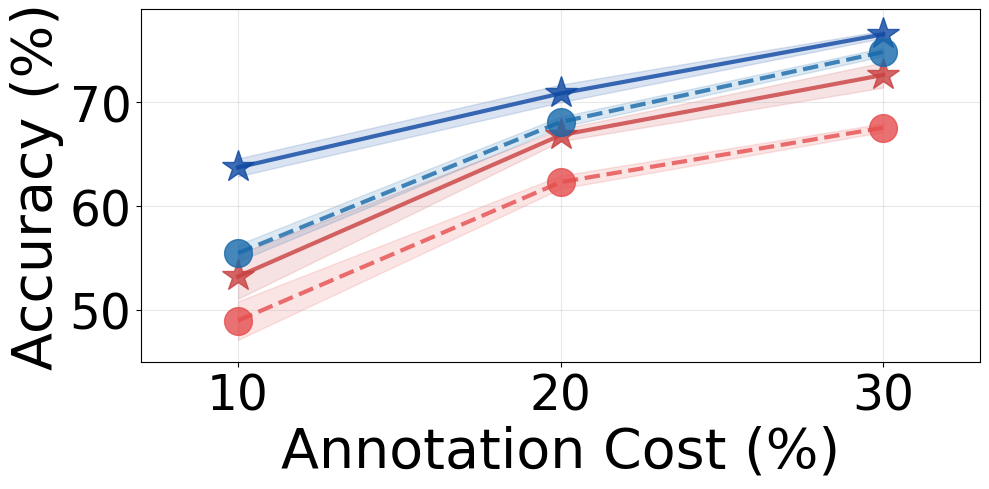}
        \vspace{-0.6cm}
        \caption{RefCOCO} 
        \label{fig:main_rec_a}
    \end{subfigure}
    \hfill 
    \begin{subfigure}{0.31\textwidth}
        \includegraphics[width=\linewidth]{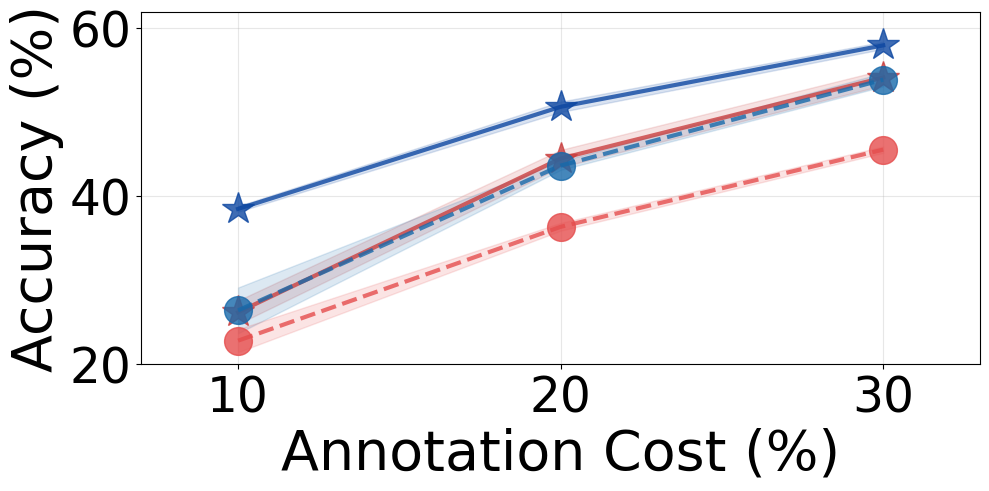}
        \vspace{-0.6cm}
        \caption{RefCOCO+} 
        \label{fig:main_rec_b}
    \end{subfigure}
    \hfill 
    \begin{subfigure}{0.31\textwidth}
        \includegraphics[width=\linewidth]{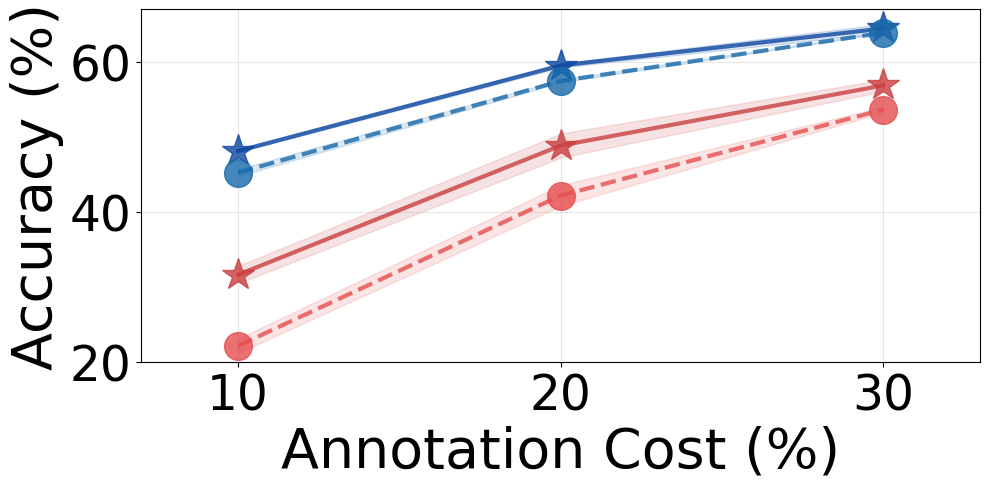}
        \vspace{-0.6cm}
        \caption{RefCOCOg} 
        \label{fig:main_rec_c}
    \end{subfigure}

        \vspace{-0.3cm}
    \caption{Comparison with the na\"ive labeling and random acquisition on a \textbf{REC} task. The reported results are averaged on the val and test (A,B) splits of each dataset. The annotation cost (x-axis) is calculated as the measured cost of the acquired samples via used labeling tool divided by the total manual annotation cost of the dataset.}
            \vspace{-0.5cm}
    \label{fig:main_rec_figure}
\end{figure*}

\begin{figure*}[t]
    \centering
    \includegraphics[width=1.2\textwidth,center]{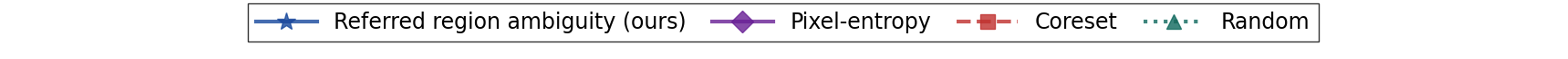}
    \vspace{-0.3cm}
    \begin{subfigure}{0.32\textwidth}
        \includegraphics[width=\linewidth]{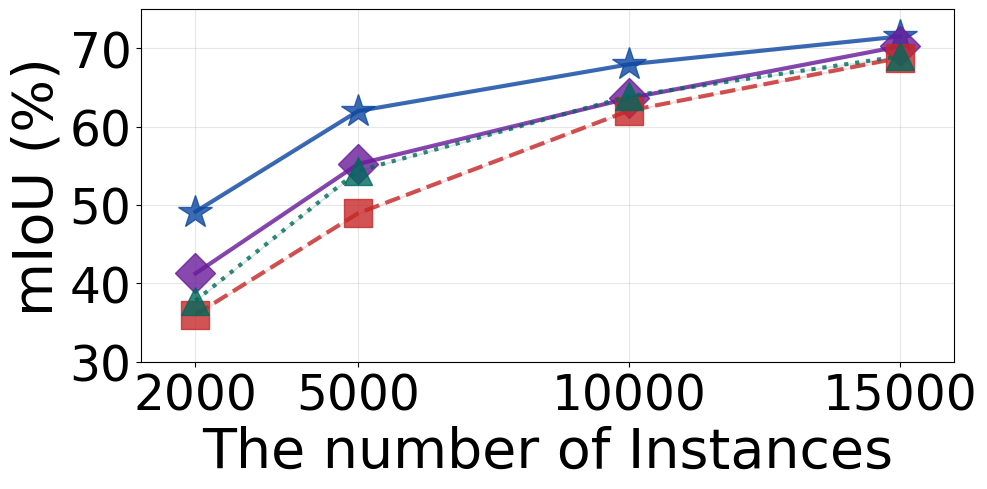}
        \vspace{-0.6cm}
        \caption{RefCOCO} 
        \label{fig:main_ris_acq_a}
    \end{subfigure}
    \hfill 
    \begin{subfigure}{0.32\textwidth}
        \includegraphics[width=\linewidth]{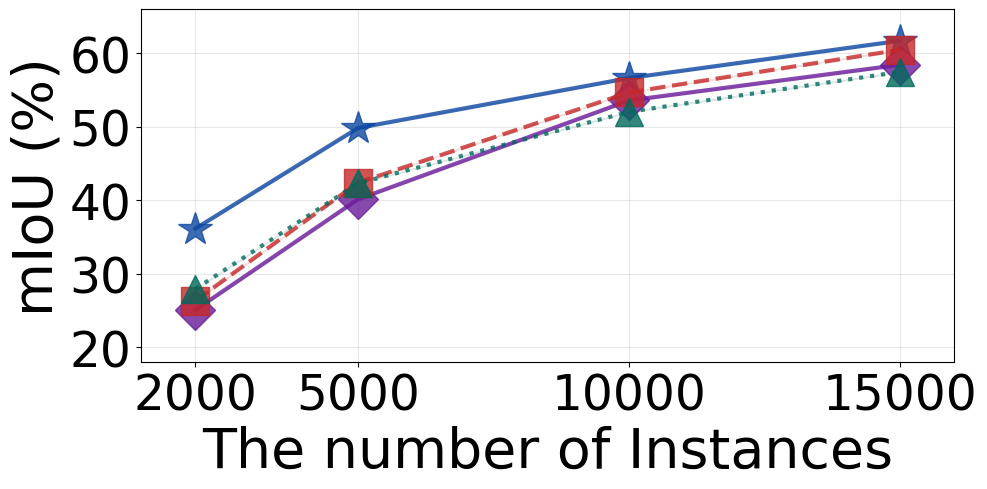}
        \vspace{-0.6cm}
        \caption{RefCOCO+} 
        \label{fig:main_ris_acq_b}
    \end{subfigure}
    \hfill 
    \begin{subfigure}{0.32\textwidth}
        \includegraphics[width=\linewidth]{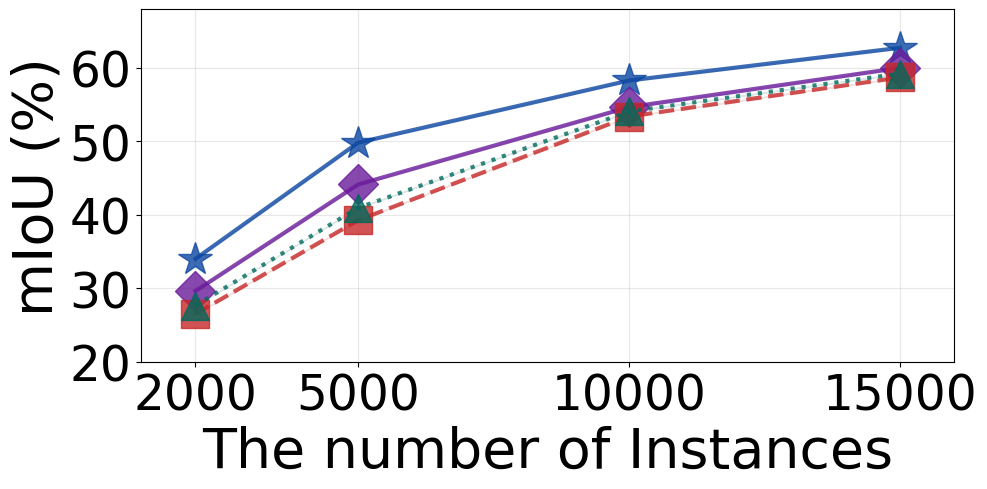}
        \vspace{-0.6cm}
        \caption{RefCOCOg} 
        \label{fig:main_ris_acq_c}
    \end{subfigure}

        \vspace{-0.05cm}
    \caption{Comparisons with several acquisition baselines on a \textbf{RIS} task.}
            \vspace{-0.3cm}
    \label{fig:main_ris_figure_acq}
\end{figure*}

\begin{figure*}[t]
    \centering 
    \vspace{0.4cm}
    \vspace{-0.3cm}
    \begin{subfigure}{0.32\textwidth}
        \includegraphics[width=\linewidth]{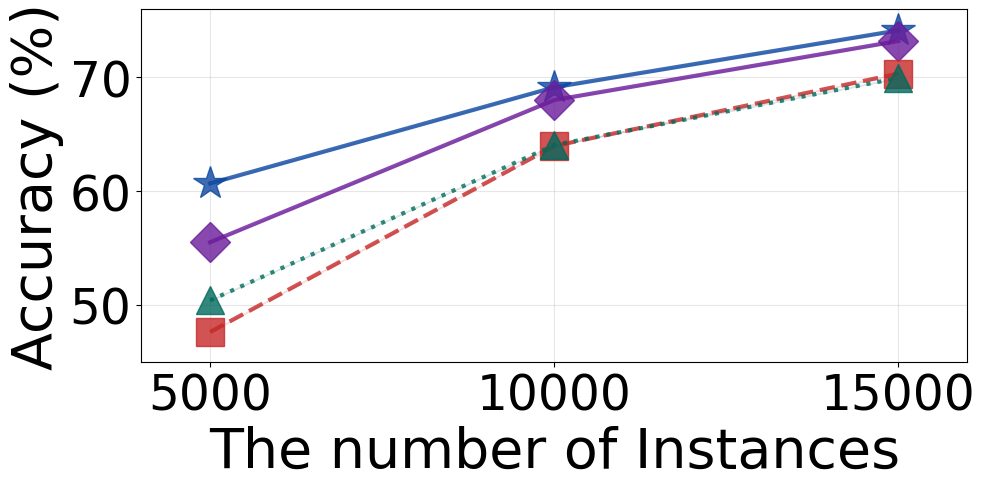}
        \vspace{-0.6cm}
        \caption{RefCOCO} 
        \label{fig:main_rec_acq_a}
    \end{subfigure}
    \hfill 
    \begin{subfigure}{0.32\textwidth}
        \includegraphics[width=\linewidth]{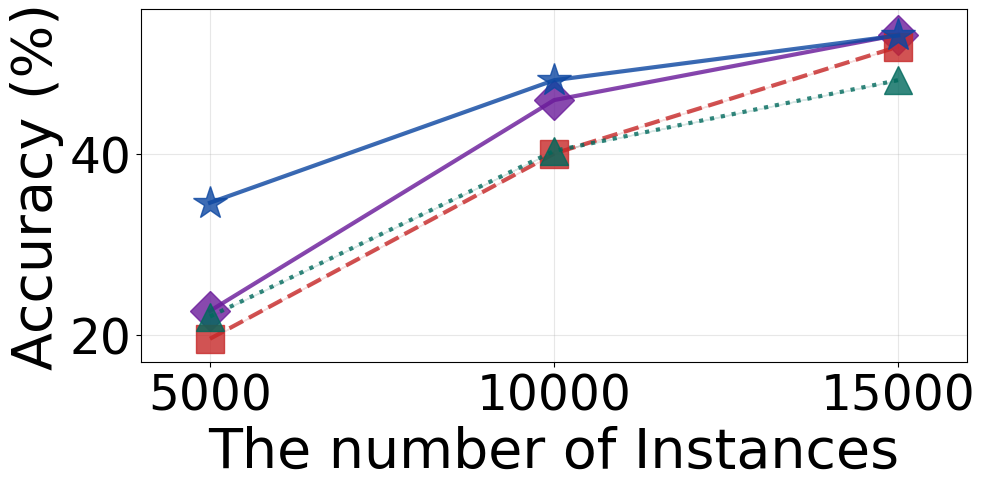}
        \vspace{-0.6cm}
        \caption{RefCOCO+} 
        \label{fig:main_rec_acq_b}
    \end{subfigure}
    \hfill 
    \begin{subfigure}{0.32\textwidth}
        \includegraphics[width=\linewidth]{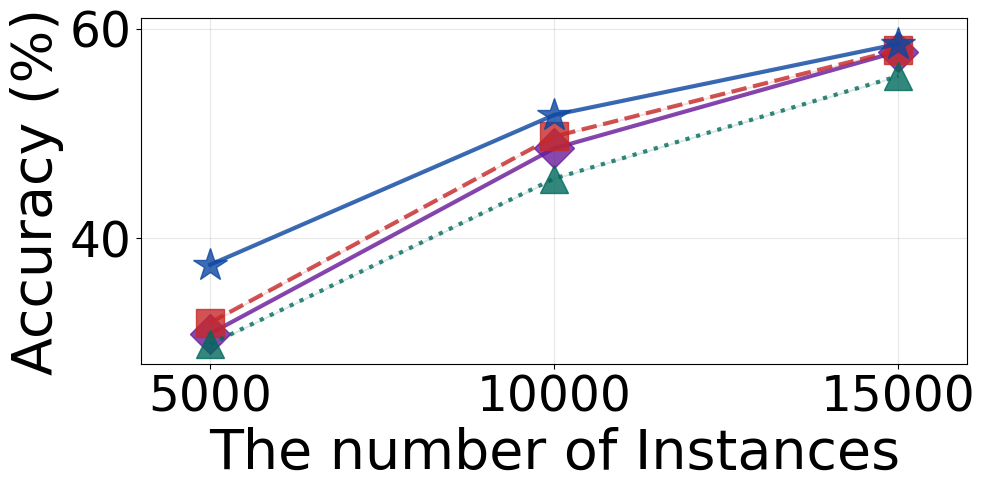}
        \vspace{-0.6cm}
        \caption{RefCOCOg} 
        \label{fig:main_rec_acq_c}
    \end{subfigure}
        \vspace{-0.3cm}
    \caption{Comparisons with several acquisition baselines on a \textbf{REC} task.    By comparing performance based on the number of acquired instance-text pairs, we isolate the effectiveness of each acquisition from the annotation tool.}
    \vspace{-0.2cm}
    \label{fig:main_rec_figure_acq}
\end{figure*}

\section{Experiments}
\vspace{-0.15cm}
\label{sec:exp}


\subsection{Implementation Details}
\vspace{-0.15cm}
\textbf{For generation of auxiliary region-text pairs} (Sec.~\ref{subsec:pseudo_data}), 
We extract potential regions from Grounded-SAM~\cite{grounded_sam} using a confidence threshold of 0.23, and limit the number of extracted ones per image to 7.
We then prompt ViP-LLaVA~\cite{vip_llava} with the instruction ``\textit{describe the object in the red box}'' to generate the corresponding expression. 
\textbf{For acquisition function} (Sec.~\ref{subsec:acquisition}), 
the background region is obtained as the remaining area after excluding all extracted instance regions.
\textbf{For cost-efficient annotation interface} (Sec.~\ref{subsec:annotation_tool}), we set the number of suggested word candidates to $k=5$.
The annotation interface runs on a single RTX 3090 GPU.
\textbf{For the experiments}, we train DETRIS-B~\cite{detris} model for RIS and MaPPER-B model~\cite{mapper} for REC on two A6000 GPUs.
In both models, we follow the training configurations as detailed in their respective models.
Following previous AL works~\cite{cold_2}, we train the initial model on 1\% random samples
to address the cold-start problem~\cite{cold_start}.
{The GPU costs incurred by the auxiliary generation and the annotation interface are in Appendix~\ref{app:gpu_cost}.}

\vspace{-0.2cm}
\subsection{Dataset and Metric}
\vspace{-0.15cm}
For dataset, we evaluate our active learning framework on standard benchmarks for visual grounding tasks (RIS and REC): RefCOCO~\cite{refcoco}, RefCOCO+~\cite{refcoco}, RefCOCOg~\cite{refcocog_google}.
Details on each dataset are in Appendix.
For evaluation metric,
we use standard metrics for both RIS and REC.
For RIS, we adopt mIoU (mean IoU), which is the average of the per-sample IoU scores.
For REC, we employ Prec@50, which is an accuracy only when IoU is over 0.5.

\vspace{-0.2cm}
\subsection{Main Results}
\vspace{-0.15cm}
\paragraph{Referring Image Segmentation.}
In Fig.~\ref{fig:main_ris_figure}, we compare the performance of our active learning framework with random acquisition and na\"ive labeling (\ie, polygon drawing for masks and manual typing for expressions) on RIS benchmark under annotation budgets of 2\%, 5\%, 10\%, 20\%, and 30\%.
The x-axis denotes the measured annotation cost of acquired samples divided by the total na\"ive labeling cost of the dataset.
For clarity, the reported mIoU results are averaged across the val and test(A/B) of each RefCOCO(+/g) and three independent runs to ensure statistical significance.
The proposed acquisition outperforms random acquisition under both labeling settings (\ie, our efficient tool and na\"ive tool), while our annotation interface delivers strong performance across all budgets, proving its efficiency.


\vspace{-0.2cm}

\begin{table*}[t!]
    \centering
    \small
    \renewcommand{\arraystretch}{1.2}
    \scalebox{0.99}{
    \begin{tabular}{ccc|ccccc|c}
    \cline{1-9}
    \multirow{2}{*}{\makecell{Acquisition\\Function}} &\multicolumn{2}{|c|}{Labeling Interface}     & \multicolumn{5}{c|}{Annotation Cost} & \multirow{2}{*}{Average} \\ \cline{2-8}
    &\multicolumn{1}{|c|}{~~Mask~~} & ~~Text~~ &2\% & 5\% & 10\% & 20\% & 30\% & \\ \cline{1-9}   
    & & & ~31.20$_{\pm1.98}$~ & ~45.04$_{\pm0.97}$~ & ~56.10$_{\pm1.44}$~ & ~66.37$_{\pm0.34}$~ & ~69.53$_{\pm0.25}$~ & 53.65 \\
    \checkmark & & & ~36.53$_{\pm0.90}$~ & ~52.05$_{\pm1.28}$~ & ~60.84$_{\pm0.56}$~ & ~68.20$_{\pm0.14}$~ & ~70.65$_{\pm0.22}$~ & 57.66\\
    \checkmark & \checkmark & & ~38.08$_{\pm0.52}$~ & ~54.09$_{\pm1.30}$~ & ~62.71$_{\pm0.49}$~ & ~69.83$_{\pm0.17}$~ & ~71.92$_{\pm0.40}$~ & 59.33 \\
    \checkmark & & \checkmark & ~38.66$_{\pm1.18}$~ & ~56.40$_{\pm1.05}$~ & ~63.42$_{\pm0.44}$~ & ~70.01$_{\pm0.24}$~ & ~72.37$_{\pm0.31}$~ & 60.17 \\
    \checkmark & \checkmark & \checkmark & ~\textbf{45.41}$_{\pm1.07}$~ & ~\textbf{58.94}$_{\pm0.86}$~ & ~\textbf{66.22}$_{\pm0.51}$~ & ~\textbf{71.74}$_{\pm0.19}$~ & ~\textbf{73.50}$_{\pm0.29}$~ & \textbf{63.16}\\
     \cline{1-9}   
    \end{tabular}
    }
    \vspace{-0.3cm}
    \caption{Ablation study within proposed components. Additional analyses are provided in Appendix~\ref{supple:additional_exp}.}
    \vspace{-0.4cm}
    \label{tab:ablation_study}
\end{table*}

\begin{table}[t]
    \centering
    \renewcommand{\arraystretch}{1.2}
    \scriptsize
    \setlength{\tabcolsep}{3.4pt} 
    \scalebox{1.08}{
    \begin{tabular}{l|cc|ccc}
    \cline{1-6}
    \multirow{2}{*}{Tools}      & \multicolumn{2}{c}{Mask Labeling} & \multicolumn{3}{|c}{Referring-Expression Labeling} \\ \cline{2-6}
    & Time (s) & mIoU & Time (s) & Sen.-BERT Sim. & CLIP Score \\
    \cline{1-6}
    Na\"ive     & 60.6 & 73.0 & 45.8 & 39.9 & 19.2\\
    Ours & 9.9 & 80.4 & 28.6 & 44.4 & 22.2 \\
    \cline{1-6}
    \end{tabular}
    }
    \vspace{-0.3cm}
    \caption{User study comparisons with the na\"ive labeling. We involve 30 annotators from Amazon Mechanical Turk. Time (s) means the average time per sample.}
    \label{tab:user_study}
    \vspace{-0.5cm}
\end{table}



\paragraph{Referring Expression Comprehension.}
We also present the performance of our method on REC compared to random acquisition and na\"ive labeling (\ie, solely manual typing for expressions) under 10\%, 20\%, and 30\% annotation budgets in Fig.~\ref{fig:main_rec_figure}.
Results are averaged over the val and test(A/B) splits of RefCOCO(+/g) and three separate runs.
Our acquisition consistently outperforms random selection, while our text annotation tool achieves higher performance than the na\"ive tool at the same annotation cost, demonstrating its cost efficiency.

\vspace{-0.2cm}

\paragraph{Comparisons on Acquisition Function.}
\label{para:comp_acq}
To isolate the effect of our acquisition, we compare it with other baselines on RIS and REC using the number of selected region-text pairs as the budget, removing the influence of labeling cost efficiency. 
We consider random selection, pixel-entropy~\cite{pixel-entropy}, and coreset~\cite{coreset}, detailed in Appendix~\ref{supple:details_acq}. 
As shown in Figs.~\ref{fig:main_ris_figure_acq} and \ref{fig:main_rec_figure_acq}, our referred region ambiguity consistently outperforms all baselines, demonstrating its effectiveness in selecting informative VG samples.


\vspace{-0.2cm}

\paragraph{Verification of Annotation Tool with User Study.}
In Tab.~\ref{tab:user_study}, we empirically validate the efficiency of our proposed labeling tool via a user study with 30 annotators on Amazon Mechanical Turk.
We compare it with the na\"ive labeling approach (\ie, polygon clicks for masks and manual typing for text), using per-sample time and quality metrics, including mIoU for masks, Sentence-BERT~\cite{sentencebert} similarity between expressions and ground-truth one, and CLIP score~\cite{clipscore} between cropped images around masks and expressions. 
Our interface achieves faster annotation while maintaining high-quality mask specification and referring-expression labeling.
Examples of user-annotated samples for each tool are visualized in Fig.~\ref{fig:supple_qual_user} of Appendix~\ref{app:subsec:user_study}.


\vspace{-0.2cm}

\subsection{Ablation Study}
\vspace{-0.1cm}
We conduct ablation studies on RefCOCO for RIS, reporting average mIoU over the val, test(A/B) splits to validate the effectiveness of our approach.

\vspace{-0.2cm}

\paragraph{Effect of Each Proposed Component.}
Tab.~\ref{tab:ablation_study} presents the contribution of each component in our AL framework.
Starting from random acquisition with na\"ive annotation, replacing random selection with our referred region ambiguity notably improves performance, especially under low budgets.
Adding efficient mask and text-labeling tools further reduces annotation costs and improves performance, with their combination yielding the best results.
Additional analyses (\eg, auxiliary regions with other class vocabularies and GPU cost) are in Appendix~\ref{supple:additional_exp}.



\vspace{-0.2cm}

\begin{table}[t]
    \centering
    \scriptsize
        \setlength{\tabcolsep}{4pt} 
    \renewcommand{\arraystretch}{1.2}
    \scalebox{1.07}{
    \begin{tabular}{@{\hspace{2pt}}l|ccc|c@{\hspace{2pt}}}
    \cline{1-5}      
    \multirow{2}{*}{Ablations}  & \multicolumn{3}{c|}{Annotation Budget} & \multirow{2}{*}{\makecell{~Measured Cost\\~over Dataset}}\\ \cline{2-4}      
                & 2\%  & 10\% & 20\%  \\ \cline{1-5}   
    \multicolumn{5}{l}{~~~(a) Mask Annotations} \\ \cdashline{1-5}[0.2pt/1pt]
    Na\"ive polygon & 38.66 & 63.42 & 70.01 & 1,616,155\\ 
    Superpixel & 43.94& 64.80  & 71.08 &  527,770\\  \cdashline{1-5}[0.2pt/1pt]
    Pre-extracted SAM (ours) & \textbf{45.41}  & \textbf{66.22}  & \textbf{71.74} & \textbf{53,970} \\ \cline{1-5}   
    \multicolumn{5}{l}{~~~(b) Text Annotations} \\ \cdashline{1-5}[0.2pt/1pt]
    Na\"ive typing & 38.08  & 62.71 & 69.83 & 6,493,300 \\
    Sentence selection & 40.18 & 64.34  & 70.16 & \textbf{2,814,811} \\  \cdashline{1-5}[0.2pt/1pt]
    Word selection (ours) & \textbf{45.41} &\textbf{66.22} & \textbf{71.74} & 4,238,677 \\
    \cline{1-5}   
    \end{tabular}
    }
    \vspace{-0.25cm}
    \caption{Ablation on the annotation interface. Measured cost denotes the total labeling cost required to label all ground-truth data in RefCOCO with each tool.
    }
    \vspace{-0.5cm}
    \label{tab:interface}
\end{table}


\paragraph{Ablation in Annotation Interface.}
To confirm the effectiveness of the proposed labeling interface, Tab.~\ref{tab:interface} compares our tool with several baseline labeling strategies.
\textbf{For mask labeling}, we compare 1) \textit{na\"ive polygon}, where the user clicks a series of vertices around region boundaries, and 2) \textit{superpixel}~\cite{seeds}, where the user selects over-segmented superpixels to produce a mask.
Compared to them, our mask tool, based on pre-extracted SAM masks, achieves the best results across all budgets and the lowest mask click cost, demonstrating its superiority.
\textbf{For text labeling}, we compare 1) \textit{manual typing}, where annotators manually write full sentences from scratch, and 2) \textit{sentence selections}, where annotators select the most suitable expression among several MLLM-generated candidates.
Our word-level selection approach with MLLM-guided decoding achieves the highest performance with the reduced manual effort over all compared methods, enabling more high-quality and faster text labeling, as validated in the user study (Tab.~\ref{tab:user_study}).


\begin{figure*}[t]
    \centering 

    \begin{subfigure}{0.48\textwidth}
        \includegraphics[width=\linewidth]{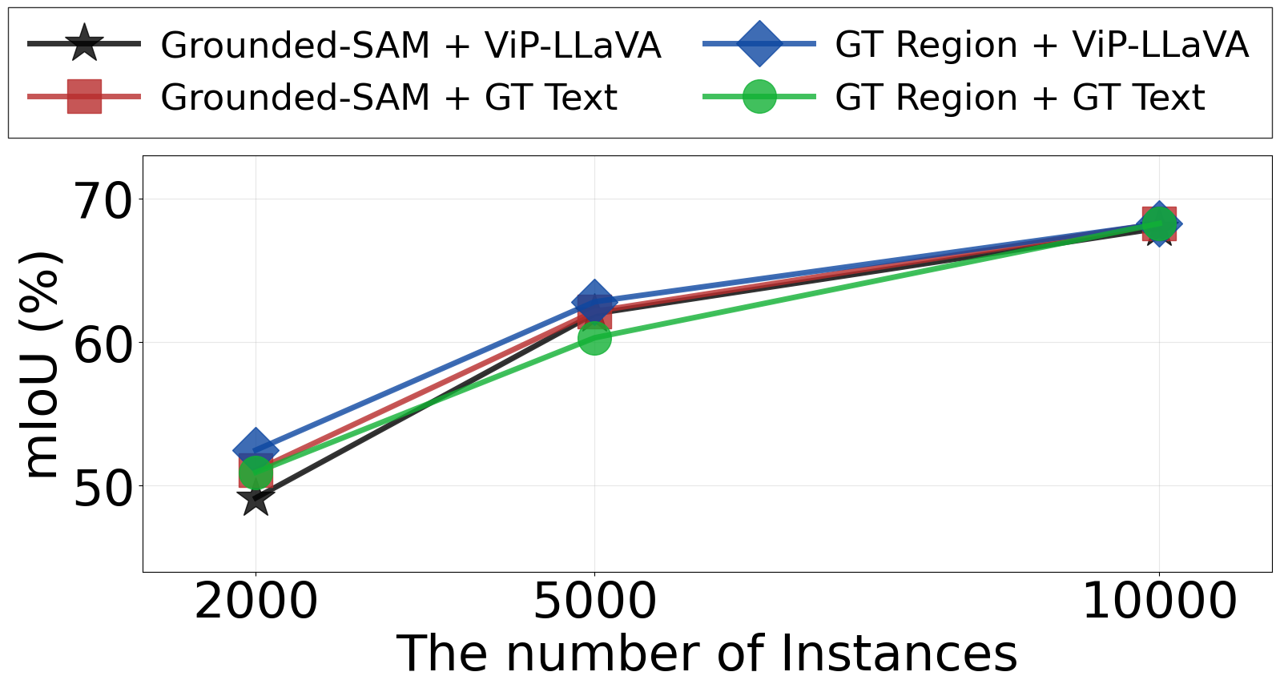}
        \caption{Ablation with GT annotations} 
        \label{fig:sub_b_aux}
    \end{subfigure}
    \hfill 
    \begin{subfigure}{0.48\textwidth}
        \includegraphics[width=\linewidth]{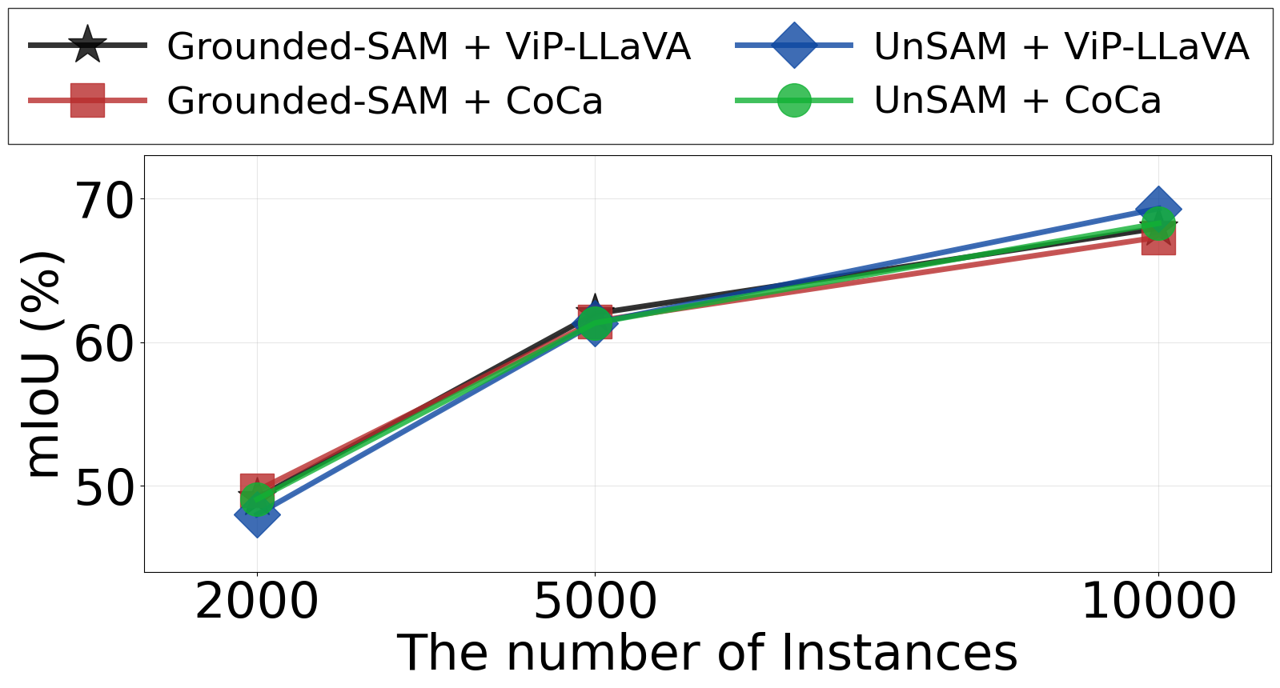}
        \caption{Ablation with other generators} 
        \label{fig:sub_c_aux}
    \end{subfigure}

        \vspace{-0.2cm}
    \caption{Ablation on the auxiliary generator. (a) we use GT labels as auxiliary pairs and (b) we use UnSAM~\cite{unsam} (unsupervised model) for region generator and CoCa~\cite{coca} (trained solely on web-crawled data without human annotations) for text generator.}
    \label{fig:auxiliary_ablation}
\end{figure*}

\begin{table*}[t!]
    \centering
    \footnotesize
    \renewcommand{\arraystretch}{1.2}
    \setlength{\tabcolsep}{8pt}
    \scalebox{1.}{
    \begin{tabular}{@{\hspace{4pt}}l|ccc|ccc@{\hspace{6pt}}}
    \cline{1-7}
    \multirow{2}{*}{~Quality~}     & \multicolumn{3}{c|}{Auxiliary Region Generator} & \multicolumn{3}{c}{Auxiliary Text Generator} \\ \cline{2-7}
         & GT mask & Grounded-SAM & UnSAM & GT text & ViP-LLaVA & CoCa \\ \cline{1-7}
    ~mIoU or Sim.~ & 100 & 83.56 & 56.85 & 100 & 54.72  & 32.97 \\ \cline{1-7}
    \end{tabular}
    }
    \vspace{-0.1cm}
    \caption{Quality of auxiliary pairs generated from the auxiliary generators in the ablation (Fig.~\ref{fig:sub_c_aux}). For the region quality, we use mIoU with the ground-truth masks. For the text quality, we use Sentence-BERT~\cite{sentencebert} similarity with the ground-truth text.}
    \vspace{-0.4cm}
    \label{tab:quality_auxiliary}
\end{table*}

\vspace{-0.1cm}

\paragraph{Ablation on Auxiliary Generator.}
{
To examine the sensitivity of our active learning framework to auxiliary generators, we conduct ablations using diverse auxiliary sources in Fig.~\ref{fig:auxiliary_ablation}.
When using the ground-truth annotations as auxiliary region-text pairs (Fig.~\ref{fig:sub_b_aux}), our AL method consistently remains robust.
For the other generators (Fig.~\ref{fig:sub_c_aux}), we use UnSAM~\cite{unsam} (unsupervised model) for region generator and CoCa~\cite{coca} (with the weights trained solely on web-crawled data without human annotations) for text generation.
Despite quality differences among auxiliary sources (Tab.~\ref{tab:quality_auxiliary}), our active learning framework consistently maintains robust performance.
These results indicate that our approach is not sensitive to the auxiliary generators.
We attribute this robustness to the role of auxiliary pairs as candidate proposals for acquisition rather than final supervision;
active learning mainly depends on the relative ranking of informative candidates, so moderate noise in masks or text often does not substantially change the priority of selection.
}

\vspace{-0.1cm}
\subsection{Qualitative Analysis}
\vspace{-0.1cm}

\begin{figure*}[t] 
    \centering
    \hfill
    \begin{subfigure}{0.32\textwidth} 
        \centering
        \includegraphics[width=\linewidth]{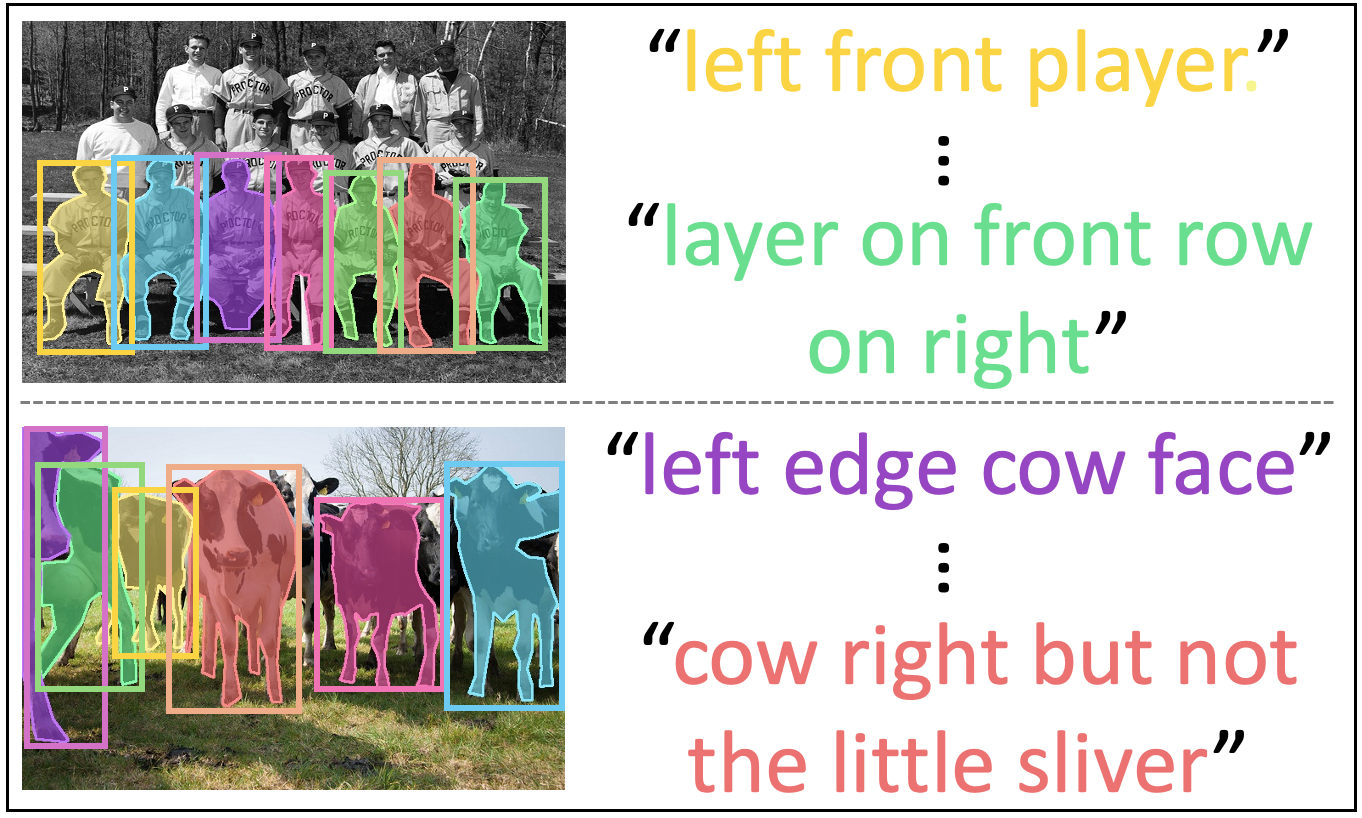}
        \vspace{-0.6cm}
        \caption{Ours}
        \label{fig:acq_qual_a} 
    \end{subfigure}
    \hfill 
    \begin{subfigure}{0.32\textwidth}
        \centering
        \includegraphics[width=\linewidth]{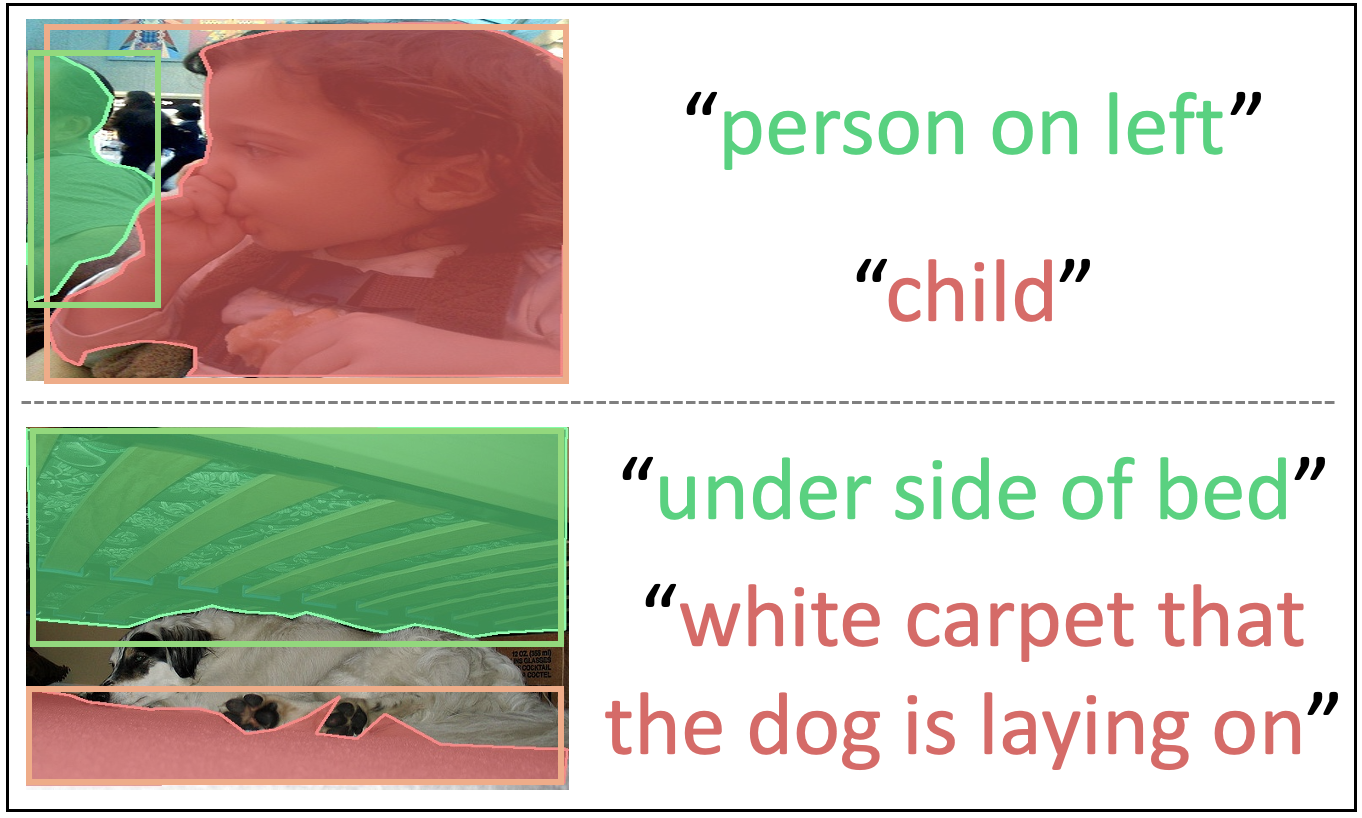}
        \vspace{-0.6cm}
        \caption{Pixel-entropy}
        \label{fig:acq_qual_b} 
    \end{subfigure}
    \hfill
    \begin{subfigure}{0.32\textwidth}
        \centering
        \includegraphics[width=\linewidth]{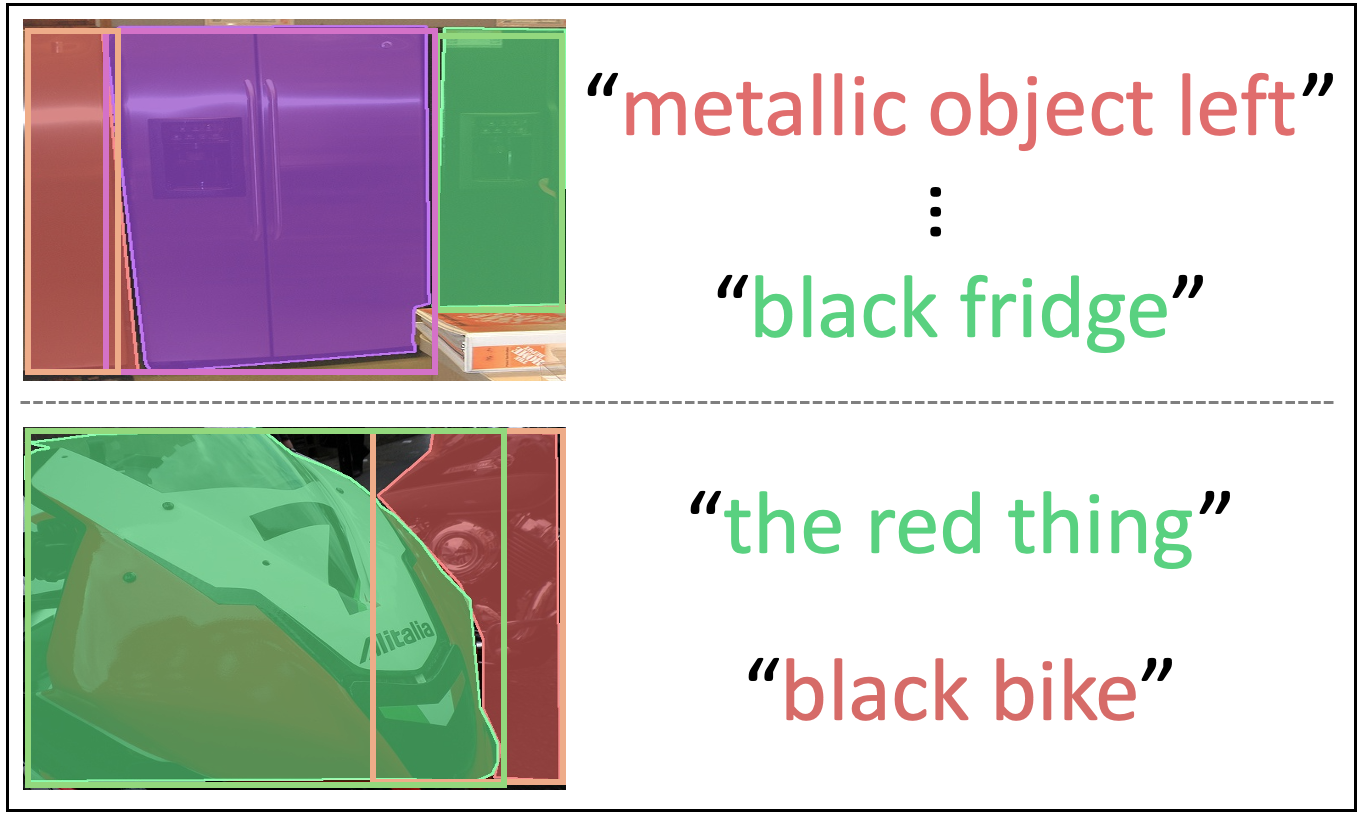}
        \vspace{-0.6cm}
        \caption{Coreset}
        \label{fig:acq_qual_c} 
    \end{subfigure}
    \hfill
    \vspace{-0.3cm}
    \caption{
    Visualization of selected images compared with other acquisition functions. 
    Paired ground-truth region-text annotations are denoted by matching colors.
    More visualization are present in Fig.~\ref{fig:supple_acq_qual} of Appendix~\ref{app:subsec:selected_samples}.
    } 
    \vspace{-0.1cm}
    \label{fig:acq_qual} 
\end{figure*}

\begin{figure*}[t]
    \centering
    \begin{minipage}[t]{0.44\linewidth}
        \centering
        \includegraphics[width=\linewidth]{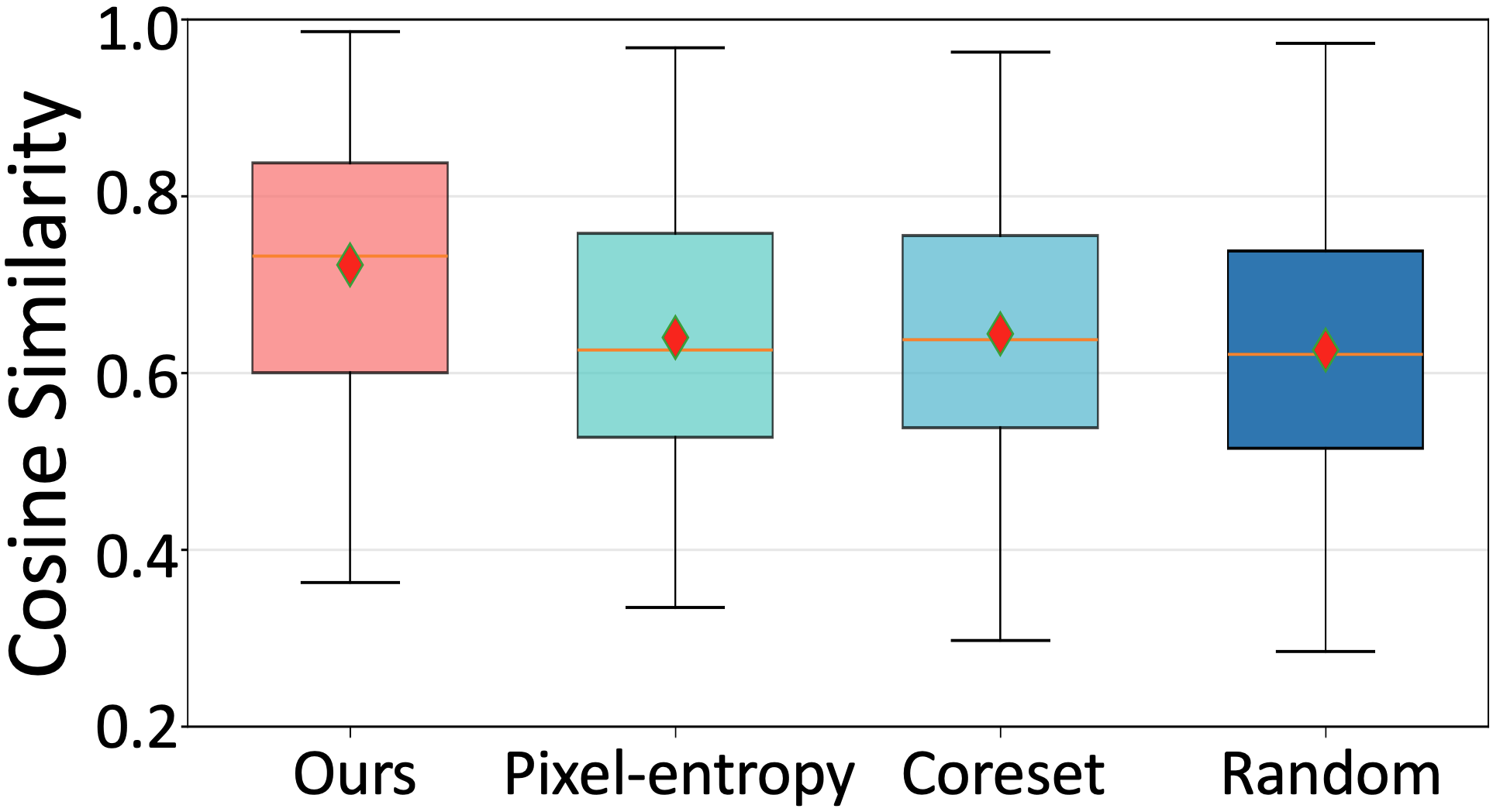}
        \vspace{-0.6cm}
        \captionof{figure}{Statistical comparison of selected pool with other methods. The measured similarity is the average feature similarity between the referred region and all other regions within the image, calculated across all samples in the selected pool.}
        \label{fig:acq_sim_scores}
    \end{minipage}
    \hfill
    \begin{minipage}[t]{0.53\linewidth}
        \centering
        \includegraphics[width=0.98\linewidth]{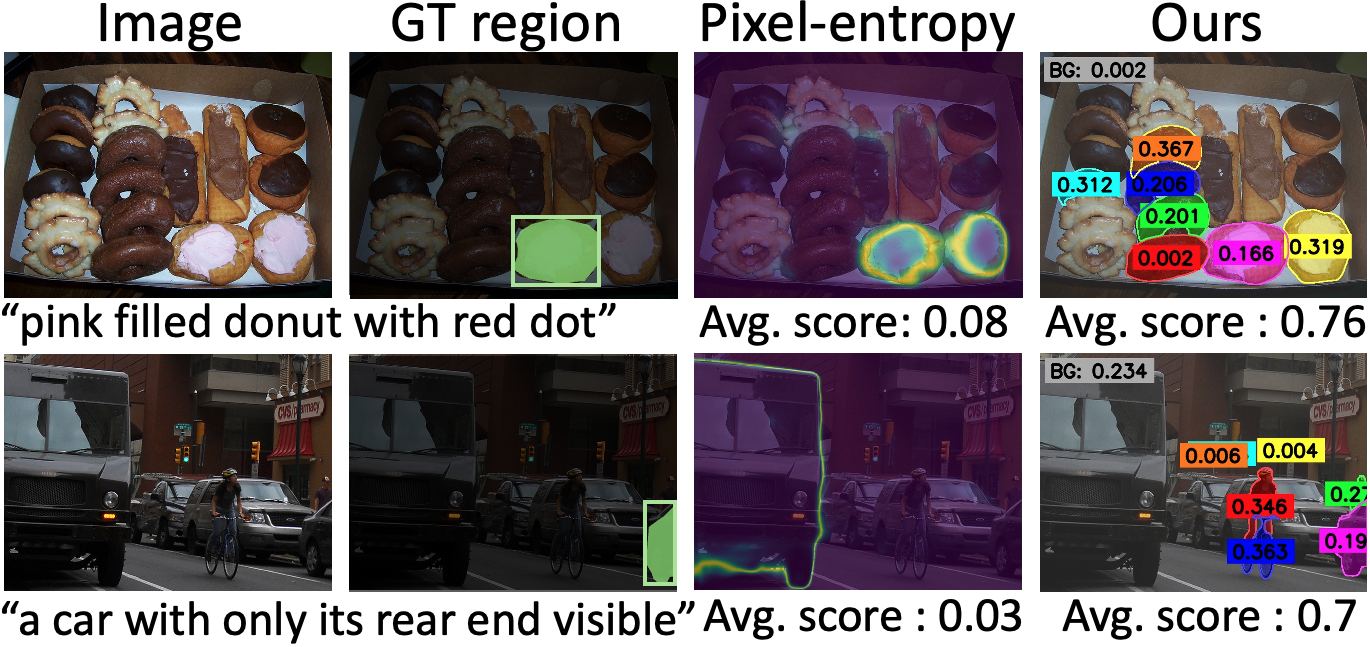}
        \vspace{-0.092cm}
        \captionof{figure}{Qualitative analysis on the process of computing the informativeness score compared to pixel-entropy. The pixel-entropy displays a raw entropy map derived from the confidence map, while ours shows the measured region-wise entropy computed from region-wise confidence distribution.}
        \label{fig:qual_score}
    \end{minipage}
    \vspace{-0.3cm}
\end{figure*}

\paragraph{Analysis on Selected Samples via Acquisition.}
To highlight that our acquisition selects more complex VG images,
we present qualitative examples of selected images in Fig.~\ref{fig:acq_qual} and a statistical analysis in Fig.~\ref{fig:acq_sim_scores} compared with pixel-entropy~\cite{pixel-entropy} and coreset~\cite{coreset}.
Our method often selects images with visual similar regions, which is a hard scenario that can confuse the model, and thus promotes its discrimination capability.
For a statistical comparison, we measure the feature similarity between the referred region and all other regions within an image, and then average this over all selected samples.
The high similarity achieved by our acquisition confirms the presence of many visually similar distractors in the selected samples, indicating that ours effectively prioritizes challenging VG samples.

\vspace{-0.1cm}

\paragraph{Analysis of Referred Region Ambiguity.}
In Fig.~\ref{fig:qual_score}, we visualize the process of computing our informativeness score and compare it with the pixel-entropy~\cite{pixel-entropy}.
Pixel-entropy estimates the informativeness by averaging the entropy values (visualized in the figure) across all pixels in the confidence map.
In contrast, our referred region ambiguity quantifies the region-level ambiguity (also visualized in the figure) from region-wise confidence scores, and then averages these region-wise entropy values.
By focusing on region-level scores, our approach effectively selects complex yet informative VG examples.
More visualization are provided in Fig.~\ref{fig:supple_qual_score} of Appendix~\ref{app:subsec:qual_score}.

\vspace{-0.15cm}
\section{Conclusion}
\label{sec:conclusion}
\vspace{-0.2cm}

In this paper, we present a novel active learning framework for visual grounding that operates under the realistic setting of a purely raw-image pool by utilizing auxiliary region-text samples as potential signals.
By estimating VG-specific informativeness via our tailored acquisition and simplifying the labor-intensive labeling process via the proposed interface, our approach significantly reduces overall manual effort while consistently outperforming all AL baselines on RIS and REC benchmarks.

\vspace{-0.2cm}

\section{Limitations}
\label{sec:limitation}
\vspace{-0.15cm}

While our annotation tool reduces human labor costs, it introduces additional GPU overhead. 
In particular, generating word-level suggestions with multimodal LLMs during annotation requires extra GPU memory and computation, as analyzed in Tab.~\ref{tab:gpu_cost} of Appendix~\ref{app:gpu_cost}. 
Since our annotation tool is designed to reduce human labeling costs, we do not explicitly consider the GPU costs incurred by the tool in its design. 
However, these costs are marginal, as the interface runs on a single RTX 3090 and requires only modest computation compared to the saved human labor. 
Overall, our tool reduces the annotation cost from \$18,296 with na\"ive labeling to \$6,752, achieving a $2.7\times$ cost reduction. 
We leave GPU-aware annotation-tool design for future work.

\vspace{-0.1cm}

\section{Ethical Consideration}
\label{sec:ethical_consideration}
\vspace{-0.15cm}
Our work aims to reduce human annotation burden in visual grounding. 
Since our annotation tool involves human annotators, they should be clearly informed of the task and fairly compensated. 
Although our interface reduces repetitive manual effort, human verification remains important to ensure accurate and unbiased referring expressions. 

\vspace{-0.15cm}

\section*{Acknowledgments}
\vspace{-0.1cm}
This work was supported by the IITP grants (RS-2019-II191906 (5\%), 2019-0-01842 (5\%), RS-2022-II220926 (35\%), RS-2026-25518317 (30\%)), the NRF grant (RS-2025-24535146 (10\%)),
the InnoCORE program (N10250156 (10\%)), GIST (Future-leading Specialized Research Project (5\%)), and the AI Computing Support Project for R\&D, funded by MSIT, Korea.


\bibliography{reference_acl}

\clearpage
\appendix
\renewcommand{\dbltopfraction}{0.9}
\renewcommand{\dblfloatpagefraction}{0.8}
\setcounter{dbltopnumber}{3} 
\raggedbottom 
\label{sec:appendix_section}

\begin{strip}
\begin{center}
   {\Large \bfseries Cost-efficient Active Learning for Referring Image Segmentation and Grounding \\
   - \textit{Appendix} - \par}
   \vspace*{24pt}
\end{center}

\vspace{-0.8cm}

\begin{center}
    \large\textbf{Table of Contents}
\end{center}

\vspace{0.1cm}

\begin{center}
\begin{minipage}{0.8\textwidth}
\begin{enumerate}[label=\Alph*., leftmargin=*, itemsep=0.2em]

    \item \hyperref[supple:additional_exp]{\textbf{Additional Experiments}} 
    \vspace{-0.1cm}
    \begin{enumerate}[label=\Alph{enumi}.\arabic*., leftmargin=2em, itemsep=0.1em]
        \item Comparisons on Ambiguous Subset
        \item Comparisons with Additional Acquisition
        \item Adaptation Results on Medical Domains
        \item Auxiliary Region with Other Class
        \item Analysis on GPU Cost
        \item Comparisons with Visual-Similarity Baselines
        \item Validity of the Bit-Based Cost Model
        \item Analysis on Collected Expressions
    \end{enumerate}
    \vspace{0.05cm}
    \item \hyperref[supple:additional_dataset]{\textbf{Additional Details}}
    \vspace{-0.1cm}
    \begin{enumerate}[label=\Alph{enumi}.\arabic*., leftmargin=2em, itemsep=0.1em]
        \item Theoretical Text Annotation Cost
        \item Acquisition Baselines
        \item User Study
        \item Dataset
    \end{enumerate}
    
    \vspace{0.05cm}
    \item \hyperref[supple:additional_qual]{\textbf{Additional Qualitative Results}} 
    \vspace{-0.1cm}
    \begin{enumerate}[label=\Alph{enumi}.\arabic*., leftmargin=2em, itemsep=0.1em]
        \item Examples of User Study Results
        \item Examples of Auxiliary Region-Text Pairs
        \item Examples on Ambiguous Subset
        \item Examples of Selected Samples via Acquisitions
        \item Examples of Referred Region Ambiguity
    \end{enumerate}

    \vspace{0.05cm}

    \vspace{0.05cm}
\end{enumerate}
\end{minipage}
\end{center}

\vspace{0.1cm}

\end{strip}

\section{Additional Experiments}
\label{supple:additional_exp}

\vspace{-0.1cm}
\begin{table*}[!b]
    \centering
    \footnotesize
    \setlength{\tabcolsep}{8pt}
    \renewcommand{\arraystretch}{1.2}
    \begin{tabular}{l|cc|cc|cc}
    \cline{1-7}
    \multirow{3}{*}{Acquisition}     & \multicolumn{6}{c}{\# of Instances} \\ \cline{2-7}
         & \multicolumn{2}{c|}{2k} & \multicolumn{2}{c|}{5k} & \multicolumn{2}{c}{10k}\\ \cline{2-7}
         & All & Subset & All & Subset & All & Subset \\ \cline{1-7}
    Random        & \textcolor{gray}{38.46} & ~32.93{\tiny\textcolor{red}{~(-14.4\%)}} &
                    \textcolor{gray}{55.64} & ~48.80{\tiny\textcolor{red}{~(-12.3\%)}} &
                    \textcolor{gray}{65.23} & 60.89{\tiny\textcolor{red}{~(-6.7\%)}} \\
    Coreset       & \textcolor{gray}{33.54} & 30.51{\tiny\textcolor{red}{~(-9.0\%)}} &
                    \textcolor{gray}{53.77} & ~43.82{\tiny\textcolor{red}{~(-18.5\%)}} &
                    \textcolor{gray}{64.33} & ~54.73{\tiny\textcolor{red}{~(-14.9\%)}} \\
    Pixel-entropy & \textcolor{gray}{39.25} & 35.74{\tiny\textcolor{red}{~(-8.9\%)}} &
                    \textcolor{gray}{56.62} & ~48.04{\tiny\textcolor{red}{~(-15.2\%)}} &
                    \textcolor{gray}{65.48} & 60.56{\tiny\textcolor{red}{~(-7.5\%)}} \\
    \cdashline{1-7}[0.2pt/1pt]
    Ours          & \textcolor{gray}{49.23} & \textbf{47.03}{\tiny\textcolor{red}{~(-4.5\%)}} &
                    \textcolor{gray}{62.91} & \textbf{58.49}{\tiny\textcolor{red}{~(-7.0\%)}} &
                    \textcolor{gray}{68.96} & \textbf{64.00}{\tiny\textcolor{red}{~(-7.2\%)}} \\
    \cline{1-7}
    \end{tabular}
    \caption{Comparisons with several acquisition baselines on an ambiguous subset of 100 highly complex images, each containing more than three objects of the same class.}
    \vspace{-0.2cm}
    \label{tab:ambiguous_subset}
\end{table*}

\subsection{Comparisons on Ambiguous Subset}
\label{supple:ambiguous_subset}
\vspace{-0.05cm}
Our acquisition function focuses on ambiguous cases, where several objects similar to the target appear in the same image.
To verify that it effectively improves the model's ability to distinguish the target from similar ones, we evaluate the RIS model trained under various acquisitions on an ambiguous subset in Tab.~\ref{tab:ambiguous_subset}.
We construct this subset from a val split of RefCOCO by selecting 100 heavily confused images that contain more than three objects of the same class (provided in the ground-truth labels). 
Our approach yields the best performance and among the smallest relative performance drops (\%) on this subset, highlighting that it indeed enhances the discriminative ability in complex scenes.
The qualitative results of predictions for each method on this set are visualized in Fig.~\ref{fig:supple_ambiguous}.

\vspace{-0.1cm}

\subsection{Comparisons with Additional Acquisition}
\label{app:additional_acquisition}
In Tab.~\ref{tab:additional_acquistions}, we compare our referred region ambiguity with additional acquisition baselines, including PPAL~\cite{ppal} and Learning Loss~\cite{lpl}, on the RefCOCO validation set for a RIS task.
To isolate the effect of acquisition from annotation tools, we evaluate all methods using the number of acquired samples as the budget.
Our acquisition achieves the best mIoU, consistent with the results in Fig.~\ref{fig:main_ris_figure_acq} of the main manuscript.
These results further demonstrate the effectiveness of our acquisition in improving the fine-grained discriminative ability required for VG.

\begin{table*}[t]
    \centering
    \footnotesize
    \setlength{\tabcolsep}{8pt}
    \renewcommand{\arraystretch}{1.2}
    \begin{tabular}{l|c|c|c}
    \cline{1-4}
    \multirow{2}{*}{Acquisition}     & \multicolumn{3}{c}{\# of Instances} \\ \cline{2-4}
         & {2k} & {5k} & {10k}\\ \cline{1-4}
    PPAL~\cite{ppal}        & {36.77}  &
                    {55.59}  &
                    {65.98}  \\
    Learning Loss~\cite{lpl}       & {39.79}  &
                    {54.82}  &
                    {64.03}  \\
    \cdashline{1-4}[0.2pt/1pt]
    Referred Region Ambiguity (ours)          & {\textbf{49.14}}  &
                    {\textbf{61.98}}  &
                    {\textbf{67.94}}  \\
    \cline{1-4}
    \end{tabular}
    \caption{Comparisons with additional acquisition baselines for a RIS task on the validation set of RefCOCO.}
    \label{tab:additional_acquistions}
\end{table*}

\begin{table*}[t]
    \centering
    \footnotesize
    \setlength{\tabcolsep}{8pt}
    \renewcommand{\arraystretch}{1.2}
    \begin{tabular}{l|c|ccc|c|ccc}
    \cline{1-9}
    \multirow{3}{*}{Acquisition}     
    & \multicolumn{4}{c|}{QaTa-COV19} 
    & \multicolumn{4}{c}{MosMedData+} \\ \cline{2-9}
    & \multicolumn{4}{c|}{\# of instances} 
    & \multicolumn{4}{c}{\# of instances}  \\ \cline{2-9}
    & 0k$^\dagger$ & 0.1k & 0.5k & 1k 
    & 0k$^\dagger$ & 0.1k & 0.3k & 0.5k \\ \cline{1-9}

    Random        
    & \multirow{4}{*}{\textcolor{gray}{12.23}} 
    & 48.72 & 60.47 & 64.09 
    & \multirow{4}{*}{\textcolor{gray}{4.52}} 
    & 39.27 & 46.89 & 49.52 \\

    Coreset        
    &   
    & 46.96 & 61.01 & 64.28 
    &  
    & 37.17 & 45.60 & 48.57 \\

    Pixel-entropy  
    & 
    & 48.95 & 59.16 & 64.22 
    &  
    & 37.10 & 46.22 & 47.54 \\

    \cdashline{1-9}[0.2pt/1pt]

    Ours           
    & 
    & \textbf{49.79} & \textbf{62.91} & \textbf{65.40} 
    &  
    & \textbf{39.68} & \textbf{47.12} & \textbf{50.26} \\

    \cline{1-9}
    \end{tabular}
    \caption{Adaptation results on medical domains compared with several acquisitions. $\dagger$ denotes the zero-shot results of the RIS model fully-trained on RefCOCO.}
    \label{tab:medical}
\end{table*}

\begin{figure}[t]
    \centering
    \vspace{0.1cm}
    \includegraphics[width=1\linewidth]{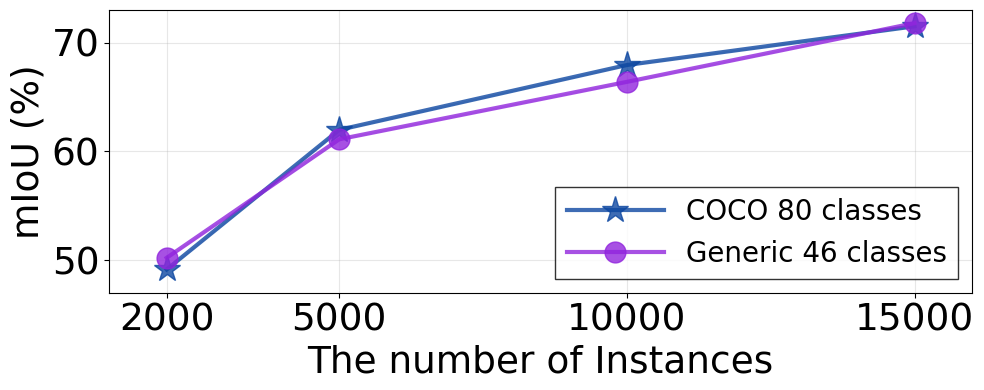}
    \vspace{-0.1cm}
    \caption{Comparison with different class sets for the auxiliary region extraction (Sec.~\ref{subsec:pseudo_data} in the manuscript) on RefCOCO dataset for a RIS task. Each class set is prompted into Grounded-SAM to extract auxiliary regions. The generic 46-class set is shown in Tab.~\ref{tab:obj365_list}.}
    \label{fig:supple_exp}
\end{figure}

\begin{table}[t]
    \centering
    \small 
    \renewcommand{\arraystretch}{1.2} 
    \scalebox{0.985}{
    \begin{tabularx}{\columnwidth}{X} 
    \toprule
    \textbf{Generic 46-class set} \\
    \midrule
    person, animal, pet, bird, mammal, sea creature, insect, 
    vehicle, car, truck, motorcycle, bicycle, aircraft, watercraft, 
    furniture, appliance, electronics, kitchenware, decoration, 
    clothing, footwear, accessories, bag, jewelry, 
    food, fruit, vegetable, meal, snack, beverage, ingredient, 
    tool, utensil, sports equipment, musical instrument, device, 
    outdoor object, sign, light, plant, building element, 
    object, container, item, product, material \\
    \bottomrule
    \end{tabularx}}
    \caption{List of a generic 46-class set, constructed by abbreviating Object365 categories using ChatGPT.}
    \label{tab:obj365_list}
\end{table}

\subsection{Adaptation Results on Medical Domains}

To examine how well our AL framework efficiently supports the domain adaptation, we fine-tune the RIS model (pretrained on RefCOCO) on medical domains with various acquisitions in Tab.~\ref{tab:medical}.
We conduct experiments on two medical referring image segmentation datasets: QaTa-COV19~\cite{qata} (5,716 train samples) and MosMedData+~\cite{mosmed} (2,183 train samples).
Due to a large domain gap (leading to poor zero-shot results), other methods may be less reliable, making random selection a competitive baseline.
Nevertheless, our acquisition exhibits faster adaptation than all the compared methods,
suggesting its effectiveness in the data-sparse domains with few samples.

\subsection{Auxiliary Region with Other Class}

In \textit{Generation of Auxiliary Region-Text Pairs} (Sec.~\ref{subsec:pseudo_data} of the main manuscript), we prompt COCO~\cite{mscoco} 80 classes into Grounded-SAM~\cite{grounded_sam} to extract auxiliary regions across all unlabeled images.
To further validate the generality of our method beyond COCO 80 categories, we conduct auxiliary region extraction with a more generic 46-class set and report the mIoU results under varying numbers of labeled instances on RefCOCO dataset (averaged over val, testA, and testB results for a RIS task) in Fig.~\ref{fig:supple_exp}.
To construct the generic 46-class vocabulary, we aggregate the fine-grained Object365~\cite{object365} categories into 46 broader superclasses using ChatGPT.
The performance gap between a COCO 80-class set and a generic 46-class set is minimal, indicating that our method is robust to the choice of object vocabulary used for generating auxiliary regions.

\subsection{Analysis on GPU Cost}

\begin{table*}[t!]
    \centering
    \footnotesize
    \setlength{\tabcolsep}{8pt}
    \renewcommand{\arraystretch}{1.2}
    \vspace{0.1cm}
    \scalebox{1}{
    \begin{tabular}{l|c|c|c||c}
    \cline{1-5}
    \multirow{2}{*}{Types}     & \multirow{2}{*}{\makecell{Generation of\\Auxiliary Pairs}} & \multirow{2}{*}{Annotation} & \multirow{2}{*}{\makecell{Total\\Cost}} & \multirow{2}{*}{\makecell{GPU\\Memory}} \\
    &  &  & \\ \cline{1-5}
    GPU overhead (w/ ours)    & 16 hours & 844 hours & \$120.4 & 8.5GB \\
    Human labor (w/ ours) & - & 844 hours & \$6,752 & - \\ \cdashline{1-5}[0.2pt/1pt]
    Human labor (w/o ours) & - & 2,287 hours & \$18,296 & -\\ 
    \cline{1-5}
    \end{tabular}
    }
    \caption{Comparisons of total annotation cost in U.S. dollars, including GPU overhead and human labor in our approach, compared with the standard baseline with na\"ive labeling and random selection. GPU cost is estimated using an RTX 3090 at \$0.14/hour on Vast.ai, and human labor cost is calculated assuming an hourly wage of \$8/hour.}
    \label{tab:gpu_cost}
\end{table*}


\begin{figure}[t]
    \centering
    \includegraphics[width=1\linewidth]{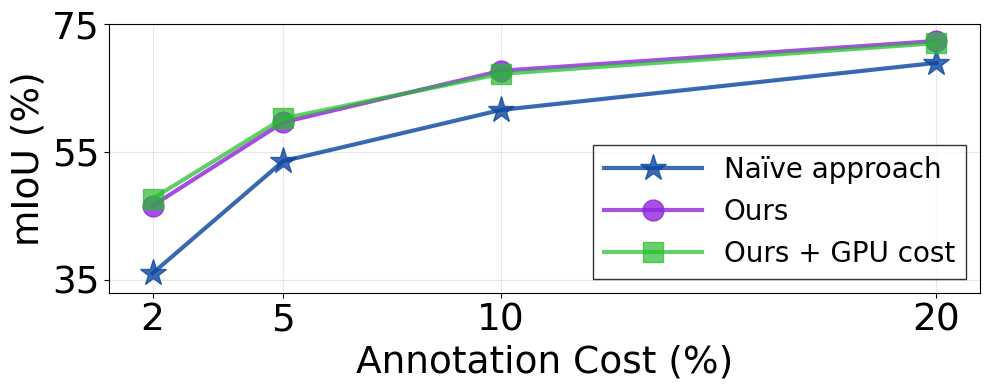}
    \vspace{-0.7cm}
    \caption{Comparisons under a fixed U.S. dollar budget that accounts for the GPU overhead in our approach. The cost estimates are summarized in Tab.~\ref{tab:gpu_cost}.}
    \label{fig:supple_gpu}
    \vspace{-0.2cm}
\end{figure}

\label{app:gpu_cost}
\vspace{-0.1cm}
The proposed AL framework introduces additional GPU overhead beyond the standard model training in two parts: (1) generating auxiliary region-text pairs using Grounded-SAM~\cite{grounded_sam} and ViP-LLaVA~\cite{vip_llava}, and (2) supporting annotation by running ViP-LLaVA in the text labeling interface for text annotation and pre-extracting SAM~\cite{sam} masks offline for the region labeling.
In this subsection, we quantify these additional GPU costs in U.S. dollars (Tab.~\ref{tab:gpu_cost}), compare them with human labor costs, and report experiments that explicitly account for resulting GPU overhead from both components (Fig.~\ref{fig:supple_gpu}).

\paragraph{GPU Overhead in Generation of Auxiliary Pairs.}
To generate auxiliary pairs, we use a single RTX 3090 GPU rented from Vast.ai at \$0.14/hour.
We run Grounded-SAM for 4 hours to produce auxiliary regions and ViP-LLaVA for 12 hours to generate auxiliary expressions, resulting in a total of 16 GPU hours.
This corresponds to an overall cost of \$2.24 (16 hours $\times$ \$0.14/hour).

\paragraph{GPU Overhead in Labeling Interface.}
For text annotation, running ViP-LLaVA in the labeling interface requires 840 GPU hours on a single RTX 3090, computed from the per-sample latency in the user study (Tab.~\ref{tab:user_study}) of the main manuscript over the entire 42,404 RefCOCO samples.
At \$0.14/hour on Vast.ai, this amounts to \$117.6.
For region annotation, we pre-extract SAM masks offline, which takes 4 GPU hours on the same GPU and costs \$0.56.
In total, the labeling interface requires 844 GPU hours, corresponding to a total GPU cost of \$118.16.
In addition, our labeling tool still requires human annotation, assuming an hourly wage of \$8, the 844 hours of human labor correspond to \$6,752, as reported in Tab.~\ref{tab:gpu_cost}.
We also report the incurred GPU memory (GB) of ViP-LLaVA-7B on an NVIDIA RTX 3090 GPU in Tab.~\ref{tab:gpu_cost}.


\paragraph{Compared to the Human Labor Costs.}
To compare these costs incurred in our approach with human manual labeling costs, we estimate the total annotation time for all 42,404 RefCOCO samples, based on the per-sample latency reported in the user study (Tab.~\ref{tab:user_study}) of the main manuscript.
The na\"ive polygon drawing for the region annotation requires 761 hours, while manually typing full sentences for the text annotation demands 1,526 hours.
In total, this amounts to 2,287 hours of human labor.
Assuming an hourly wage of \$8, the total human labor cost is \$18,296, which is substantially higher than the GPU cost incurred by our labeling interface and auxiliary region-text generation.
In Fig.~\ref{fig:supple_gpu}, we further report results under a fixed U.S. dollar budget (\%) that accounts for GPU costs, comparing na\"ive labeling with random acquisition against our labeling with our acquisition.
Despite the additional GPU overhead in our approach (Tab.~\ref{tab:gpu_cost}), the results in Fig.~\ref{fig:supple_gpu} show that our method remains more cost-effective under the same budget.
This is because the additional GPU cost is minimal compared with the human labor cost required for large-scale manual annotation.

\vspace{-0.1cm}

\begin{table*}[t]
    \centering
    \small
    \setlength{\tabcolsep}{14pt}
    \renewcommand{\arraystretch}{1.2}
    \begin{tabular}{l|c|c|c}
    \cline{1-4}
    \multirow{2}{*}{Acquisition}     & \multicolumn{3}{c}{\# of Instances} \\ \cline{2-4}
         & {2k} & {5k} & {10k}\\ \cline{1-4}
    Random                           & 37.79 & 54.36 & 63.87 \\
    DINOv2 region similarity         & 46.20 & 56.00 & 65.36 \\
    Max-region selection             & 46.32 & 59.99 & 66.86 \\
    \cdashline{1-4}[0.2pt/1pt]
    Referred Region Ambiguity (ours) & \textbf{49.14} & \textbf{61.98} & \textbf{67.94} \\
    \cline{1-4}
    \end{tabular}
    \vspace{-0.1cm}
    \caption{Comparisons with acquisition baselines that rely only on visual similarity or object density, for a RIS task on the validation set of RefCOCO.}
    \label{tab:visual_similarity}
\end{table*}

\begin{table*}[t]
    \centering
    \small
    \setlength{\tabcolsep}{12pt}
    \renewcommand{\arraystretch}{1.2}
    \begin{tabular}{l|cc|cc}
    \cline{1-5}
    \multirow{2}{*}{Interface} & \multicolumn{2}{c|}{Modeled Cost} & \multicolumn{2}{c}{Measured Time} \\ \cline{2-5}
         & Bits & Reduction & Time (s) & Reduction \\ \cline{1-5}
    Na\"ive typing         & 6,493,300 & 0.0\%  & 45.82 & 0.0\%  \\
    Sentence selection     & 2,814,811 & 56.7\% & 21.10 & 53.9\% \\
    Word selection (ours)  & 4,238,677 & 34.7\% & 28.61 & 37.6\% \\
    \cline{1-5}
    \end{tabular}
    \vspace{-0.1cm}
    \caption{Validation of the bit-based text annotation cost. We compare the reduction in the modeled bit cost with the reduction in the annotation time measured in user studies, both relative to na\"ive typing.}
    \label{tab:cost_model_validity}
\end{table*}

\subsection{Comparisons with Visual-Similarity Baselines}
\label{app:visual_similarity}
To examine whether the gains of our acquisition arise primarily from selecting images that contain many visually similar objects, we compare referred region ambiguity with two additional baselines that rely only on visual signals.
For \textit{DINOv2 region similarity}, we extract DINOv2~\cite{dinov2} features for all candidate regions in an image and use their average pairwise cosine similarity as the acquisition score.
For \textit{max-region selection}, we rank images by the number of regions proposed by Grounded-SAM~\cite{grounded_sam}, prioritizing images with more candidate objects.
Tab.~\ref{tab:visual_similarity} reports the results on the RefCOCO validation set for a RIS task, following the protocol of Appendix~\ref{app:additional_acquisition}.
Both baselines outperform random selection, suggesting that visual similarity and object density are useful acquisition signals.
However, they consistently underperform referred region ambiguity across all annotation budgets.
These results indicate that our gains cannot be explained solely by selecting images with many visually similar objects.
Rather, our criterion captures language-conditioned, model-specific uncertainty, \ie, cases in which the current grounding model cannot reliably distinguish the referred target from similar candidates given the expression.

\subsection{Validity of the Bit-Based Cost Model}
\label{app:cost_model_validity}
To further examine the validity of the bit-based text annotation cost (Eq.~\eqref{costmeasure} in Sec.~\ref{subsec:annotation_tool} of the main manuscript), we conduct an additional user study for the sentence-selection interface in Tab.~\ref{tab:interface} on Amazon Mechanical Turk.
Twenty annotators each label 10 samples, yielding 200 annotations in total.
We compare the reduction in the modeled bit cost with the reduction in the measured annotation time, both relative to na\"ive typing.
As shown in Tab.~\ref{tab:cost_model_validity}, the modeled bit-cost reduction of sentence selection closely aligns with the measured time reduction (56.7\% vs. 53.9\%), and a similar agreement is observed for our word-selection interface (34.7\% vs. 37.6\%).
Although the model cannot capture every aspect of human effort, these consistent trends across both interfaces provide additional evidence that it is a reasonable proxy for the relative annotation cost.

\subsection{Analysis on Collected Expressions}
\label{app:language_analysis}
Beyond the annotation time and similarity metrics in Tab.~\ref{tab:user_study}, we further analyze the referring expressions collected with na\"ive typing and with our interface in the user study.
Tab.~\ref{tab:language_analysis} reports the following statistics.
We use ChatGPT to categorize the words in each expression as attributes, spatial relations, or object categories, and report their proportions.
We measure diversity using the Vendi Score~\cite{vendi}, which quantifies the semantic diversity of the collected expressions based on LLM embeddings.
To assess ambiguity, we conduct a paired comparison in which ChatGPT identifies the more ambiguous expression between those produced by na\"ive typing and our interface.
For human-rated quality, we conduct an additional user study with 90 annotators from Amazon Mechanical Turk, where each annotator evaluates 20 expressions and rates their quality on a five-point scale.
To assess referring accuracy, we ask ChatGPT to identify which expression more accurately refers to the target object.

\begin{table*}[t]
    \centering
    \footnotesize
    \setlength{\tabcolsep}{5.5pt}
    \renewcommand{\arraystretch}{1.2}
    \begin{tabular}{l|c|ccc|c|c|c|c}
    \cline{1-9}
    \multirow{2}{*}{Tool} & \multirow{2}{*}{Avg. Len.} & \multicolumn{3}{c|}{Word Type} & \multirow{2}{*}{Diversity $\uparrow$} & \multirow{2}{*}{Ambiguity $\downarrow$} & \multirow{2}{*}{Quality $\uparrow$} & \multirow{2}{*}{Refer. Acc. $\uparrow$} \\ \cline{3-5}
     & & Attribute & Spatial & Category & & & & \\ \cline{1-9}
    Na\"ive typing & 12.0 & 23.6\% & 15.8\% & 9.8\%  & \textbf{20.33} & 55.5\% & 3.88 & 34.70\% \\
    Ours           & 6.2  & 23.2\% & 17.4\% & 17.4\% & 17.71 & \textbf{44.5\%} & \textbf{3.95} & \textbf{65.30\%} \\
    \cline{1-9}
    \end{tabular}
    \vspace{-0.1cm}
    \caption{Language-side analysis of the referring expressions collected with na\"ive typing and with our interface. Avg. Len. is the average number of words. Word Type is the proportion of attribute, spatial-relation, and object-category words. Diversity is the Vendi Score~\cite{vendi}. Ambiguity is the ratio of being judged as the more ambiguous expression in a paired comparison. Quality is a human-rated five-point score from 90 annotators. Refer. Acc. is the ratio of being judged as the expression that more accurately refers to the target.}
    \label{tab:language_analysis}
\end{table*}

Na\"ive typing yields a higher diversity score, likely because its expressions are longer and contain more concepts.
However, our manual inspection shows that some of these annotations describe the overall image rather than the target region.
For example, one annotation states: \textit{``Three ladies are seated close together on a sofa; one is using headphones to unwind, another is wearing a robe and face mask and using a laptop to examine her fingernails, and a fourth figure is depicted as a blue, overlaid silhouette sitting on the right.''}
This expression describes the overall scene and includes inconsistent details rather than clearly identifying the target region.
Such errors may increase diversity but also reduce clarity and increase ambiguity.
In contrast, our interface produces clearer and more target-focused expressions, as reflected in the lower ambiguity, higher human-rated quality, and substantially higher referring accuracy.

\section{Additional Details}
\label{supple:additional_dataset}

\vspace{-0.1cm}

\subsection{Theoretical Text Annotation Cost}
\label{supple:details_text_cost}
In our text annotation cost measurement (Eq.~\eqref{costmeasure} in the main manuscript, rewritten below for convenience), we rely on two types of probabilities:
(1) the probability $p_k$ that a user-desired word appears among the suggested candidates in the annotation interface (\ie, top-$k$ coverage probability), and (2) the conditional probability of a letter given the previous two letters, $p_l(l_i|l_{i-1},l_{i-2})$ (\ie, trigram probability).
In this subsection, we provide detailed statistics and empirical measurements for these two probabilities.
\begin{align*}
    C_{\text{word}} = &\underbrace{\log_2(k+1)}_{\text{Click a word or typing option}}+ \underbrace{(1-p_k)C_{\text{type}}}_{\text{Type a word manually}}, \nonumber \\
    \text{whe}\text{re}~~&C_{\text{type}} = -\sum_{i=1}^n\log_2p_l(l_i|l_{i-1},l_{i-2}).
\end{align*}

\paragraph{Top-$k$ Suggestions in Text Annotation Tool.}
To estimate top-$k$ coverage probability $p_k$, we measure the top-$k$ accuracy of the $k$-word suggestions produced by our text annotation tool on the RefCOCO dataset.
At each step of the annotation process, we mark the prediction as correct if the ground-truth next word appears within the top-$k$ suggested candidates. 
The ratio of correct predictions across all steps yields the observed top-$k$ accuracy, which we use as $p_k$.
As shown in Tab.~\ref{suptab:statistic}, $p_k$ naturally increases as $k$ grows.
However, the overall measured cost still increases because selecting from a larger candidate set incurs a higher option cost, represented by the $\log_2(k+1)$ term.
We choose $k=5$ as it provides a more intuitive word selection while achieving the lowest measured annotation cost.
\begin{table*}[t]
    \centering
    \small
    \setlength{\tabcolsep}{8pt}
    \renewcommand{\arraystretch}{1.4}
    \scalebox{1}{
    \begin{tabular}{l|cccc}
    \cline{1-5}
    \multirow{2}{*}{Metric}     & \multicolumn{4}{c}{Top-$k$ candidates} \\ \cline{2-5}
        & 3 & \textbf{5} (ours)  & 10 & 20 \\     \cline{1-5}
    Accuracy (\%) & 56.48  & 61.75 & 67.10 & ~\textbf{71.21}   \\ \cline{1-5}
    \makecell{Measured Cost\\over Dataset} & 4,274,759 & ~\textbf{4,238,677} & 4,324,874 & 4,524,190 \\ \cline{1-5}
    \end{tabular}
    }
    \caption{Text annotation costs and accuracy across varying $k$ suggested word candidates in our text annotation interface for estimating $p_k$, which represents the top-$k$ accuracy of including the ground-truth next word among the suggested $k$ candidates over the RefCOCO dataset.
    }
    \label{suptab:statistic}
\end{table*}

\paragraph{Conditional Probability of Letters.}
For the trigram conditional probability of a letter, $p_l(l_i|l_{i-1},l_{i-2})$ in Eq.~\eqref{costmeasure} of text cost measurement, we use letter-level 3-gram frequencies measured from approximately 4.5 billion characters of English text (we use the English letter frequency data presented in~\cite{letterfreq}).
For the first and second letters of a sequence, we use 1-gram and 2-gram frequencies, respectively.
Subsets of these letter frequencies are shown in Tab.~\ref{tab:ngram_stats}.
\begin{table*}[t]
    \centering
    \footnotesize  
        \setlength{\tabcolsep}{6pt}
    
    \vspace{0.2cm}
    \begin{tabular}{c cc cc cc}
        \toprule
        & \multicolumn{2}{c}{\textbf{1-gram}} & \multicolumn{2}{c}{\textbf{2-gram}} & \multicolumn{2}{c}{\textbf{3-gram}} \\
        \cmidrule(lr){2-3} \cmidrule(lr){4-5} \cmidrule(lr){6-7}
        Rank & Letters & Probability & Letters & Probability & Letters & Probability \\
        \midrule
        1  & E & 0.1210 & TH & 0.0272 & THE & 0.0181 \\
        2  & T & 0.0894 & HE & 0.0234 & AND & 0.0073 \\
        3  & A & 0.0855 & IN & 0.0204 & ING & 0.0072 \\
        4  & O & 0.0747 & ER & 0.0179 & ENT & 0.0042 \\
        5  & I & 0.0733 & AN & 0.0162 & ION & 0.0042 \\
        6  & N & 0.0717 & RE & 0.0142 & HER & 0.0036 \\
        7  & S & 0.0673 & ES & 0.0133 & FOR & 0.0034 \\
        8  & R & 0.0633 & ON & 0.0132 & THA & 0.0033 \\
        9  & H & 0.0496 & ST & 0.0126 & NTH & 0.0033 \\
        10 & L & 0.0421 & NT & 0.0118 & INT & 0.0032 \\
        \bottomrule
    \end{tabular}
    \vspace{0.2cm}
    \caption{Top-10 most frequent letter $n$-grams with their probabilities. 
    We use the English letter frequency data presented in~\cite{letterfreq}.
    }
    \label{tab:ngram_stats}
    \vspace{-0.2cm}
\end{table*}

\subsection{Acquisition Baselines}
\label{supple:details_acq}
We conduct acquisition comparisons with pixel entropy~\cite{pixel-entropy} and coreset~\cite{coreset} in Fig.~\ref{fig:main_ris_figure_acq} (RIS) and Fig.~\ref{fig:main_rec_figure_acq} (REC) of the main manuscript.
In this subsection, we provide a detailed implementation for each of the compared baselines.

\paragraph{Pixel Entropy.}
This method is a conventional acquisition function widely used in active learning for segmentation~\cite{pixel-entropy, pixel-entropy-2, pixel-entropy-viewal}.
It quantifies the informativeness of an auxiliary region-text pair by measuring how uncertain the model's prediction is at the pixel level, and it selects region-level samples for human annotations instead of image-level in ours.
Pixel entropy outputs high informativeness when the model's confidence map is uniformly spread across the whole image with moderate values (\eg, probability near 0.5).
Conversely, if the model confidently activates the whole image with the lowest or highest values (\ie, probability near 0.0 or 1.0), it outputs low informativeness.
We adapt this method for both RIS and REC models, as follows.

 \textbf{1) RIS model:} Given a specific auxiliary pair ($r^{\text{aux}},t^{\text{aux}}$), we obtain a confidence map (\ie, a final logit scoring map before thresholding), $p^i = \theta(I, t^{\text{aux}}) \in \mathbb{R}^{h \times w}$.
We then apply a sigmoid function to convert these logits into a pixel-level probability map, as follows:
$\hat p = \sigma(p) \in[0,1]^{h\times w}$.
The binary entropy $\hat H_{(x,y)}$ at each pixel ($x,y$) is defined as:
\begin{align}
    \hat H_{(x,y)} =& - \hat p_{(x,y)} \log(\hat p_{(x,y)}) \nonumber \\
    &-(1-\hat p_{(x,y)})\log(1-\hat p_{(x,y)}),
\end{align}
where $\hat H_{(x,y)}$ is the binary entropy at pixel coordinate ($x,y$) conditioned on a given auxiliary caption $t^{\text{aux}}$.
The final informativeness score $E(r^{\text{aux}},t^{\text{aux}})$ for an auxiliary pair
is obtained by averaging this pixel-level entropy $\hat H_{(x,y)}$ across all $h\times w$ pixels, as follows:
\begin{align}
    E(r^{\text{aux}},t^{\text{aux}})& = \frac{1}{h\times w}\sum_{x=1}^h\sum_{y=1}^w\hat H^i_{(x,y)}.
\end{align}
Based on this $E(r^{\text{aux}},t^{\text{aux}})$ score computed for all auxiliary pairs, we select the top-scoring auxiliary region-text pairs and annotate them at the region-level sample.

 \textbf{2) REC model:} 
Unlike RIS models that directly generate segmentation masks from a probability map, REC models predict bounding boxes by passing only the $[\text{CLS}]$ token through an MLP layer, making pixel-level entropy inapplicable. 
Instead, we leverage the attention map ${A}\in \mathbb{R}^{h \times w}$ extracted from the final decoder layer, conditioned on the given auxiliary caption $t^{\text{aux}}$ and image $I$.
We compute the normalized entropy $H_n$ of this attention map and then average across all $N$ auxiliary captions, as follows:
\begin{align}
    H_n({A})=-&\frac{\sum_{x=1}^{h}\sum_{y=1}^{w} {A}_{(x,y)} \log({A}_{(x,y)})}{\log(h \times w)},
\end{align}
where ${A}_{(x,y)}$ is an attention score at a pixel coordinate ($x,y$).
We use this $H_n(A)$ as a pixel-entropy score for each auxiliary pair in a REC model.
We then select and annotate the auxiliary pairs with the highest $H_n(A)$ scores sequentially.

\paragraph{Coreset.}
This method aims to select diverse samples from an unlabeled pool to maximize coverage of acquired samples.
To achieve this, it first extracts feature representations for all auxiliary pairs in the unlabeled auxiliary region-text pair set $\mathcal{D}_{\mathcal{U}}$ and then iteratively selects the auxiliary pair whose feature is farthest from the features of all currently labeled region-text pair set $\mathcal{D}_\mathcal{L}$.

Formally, given a specific auxiliary caption $t^{\text{aux}}$ with its image (for a labeled case, we use a ground-truth expression instead of an auxiliary caption), we extract the feature representation of a given auxiliary caption from the model encoder $\theta_{\text{enc}}$, as follows:
\begin{align}
    {F}=\theta_{\text{enc}}(I,t^{\text{aux}})\in \mathbb{R}^{h\times w\times d}, 
\end{align}
where $h,w$ are the size of the feature map and $d$ is a feature dimension.
The resulting  feature map ${F}$ is then aggregated via spatial pooling and $\ell_2$-normalized to obtain the final image feature vector $ {\hat Z}$:
\begin{align}
    {\hat Z}=\frac{{Z}}{\|{Z}\|}, ~\text{where}~ {Z} =\frac{\sum_{x=1}^h\sum_{y=1}^w { F}_{(x,y)}}{h \times w} \in \mathbb{R}^{d}, 
\end{align}
and ${F}_{(x,y)}$ denotes the feature at a pixel coordinate ($x,y$).
This normalized feature vector ${\hat Z}$ represents the auxiliary pair feature and is used for distance calculation.
For each unlabeled auxiliary pair, we compute its minimum distance over all labeled region-text pairs in the acquired pair set $\mathcal{D}_{\mathcal{L}}$:
\begin{align}
    d_j = \min_{k\in\mathcal{D}_{\mathcal{L}}} \|{\hat Z}_j - {\hat Z}_k\|_2,
\end{align}
where $k$ denotes the index for the labeled pairs.
We select the auxiliary pair with the maximum distance to the current labeled set:
\begin{equation}
    {m}^* = \arg\max_{j\in \mathcal{D}_{\mathcal{U}}} d_j,
\end{equation}
where ${m}^*$ is a selected index over all auxiliary pairs in the unlabeled pairs set $\mathcal{D}_{\mathcal{U}}$.
We add it to the labeled set, $\mathcal{D}_{\mathcal{L}} \leftarrow \mathcal{D}_{\mathcal{L}} \cup \{(r^{\text{aux}}_{{m}^*},t^{\text{aux}}_{{m}^*})\}$ with its annotations and an image.
We then update the distances for all remaining auxiliary pairs: $d_j \leftarrow \min(d_j, \|{\hat Z}_j-{\hat Z}_{{m}^*}\|), ~\forall j\in \{1,\dots,|\mathcal{D}_{\mathcal{U}}|\} \setminus \{{m}^*\}$.
This process iteratively continues until the round budget is exhausted.

\subsection{User Study}
\begin{figure*}[t]
    \centering
    \includegraphics[width=0.98\linewidth]{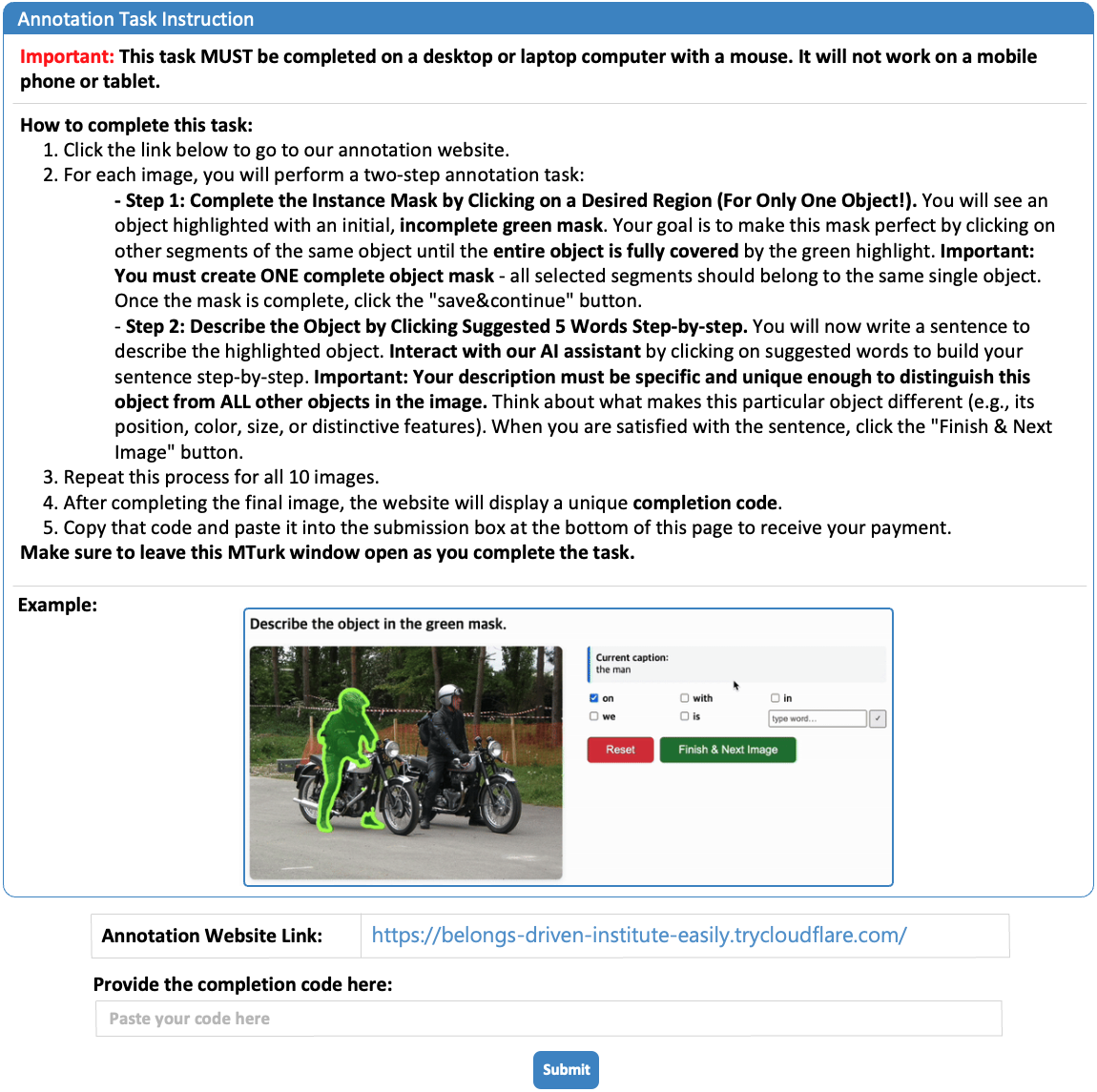}
    \caption{Annotation task instructions for the user study on Amazon Mechanical Turk.}
    \label{supfig:annotation_instructions}
\end{figure*}

\begin{figure*}[t]
    \centering
    \includegraphics[width=0.98\linewidth]{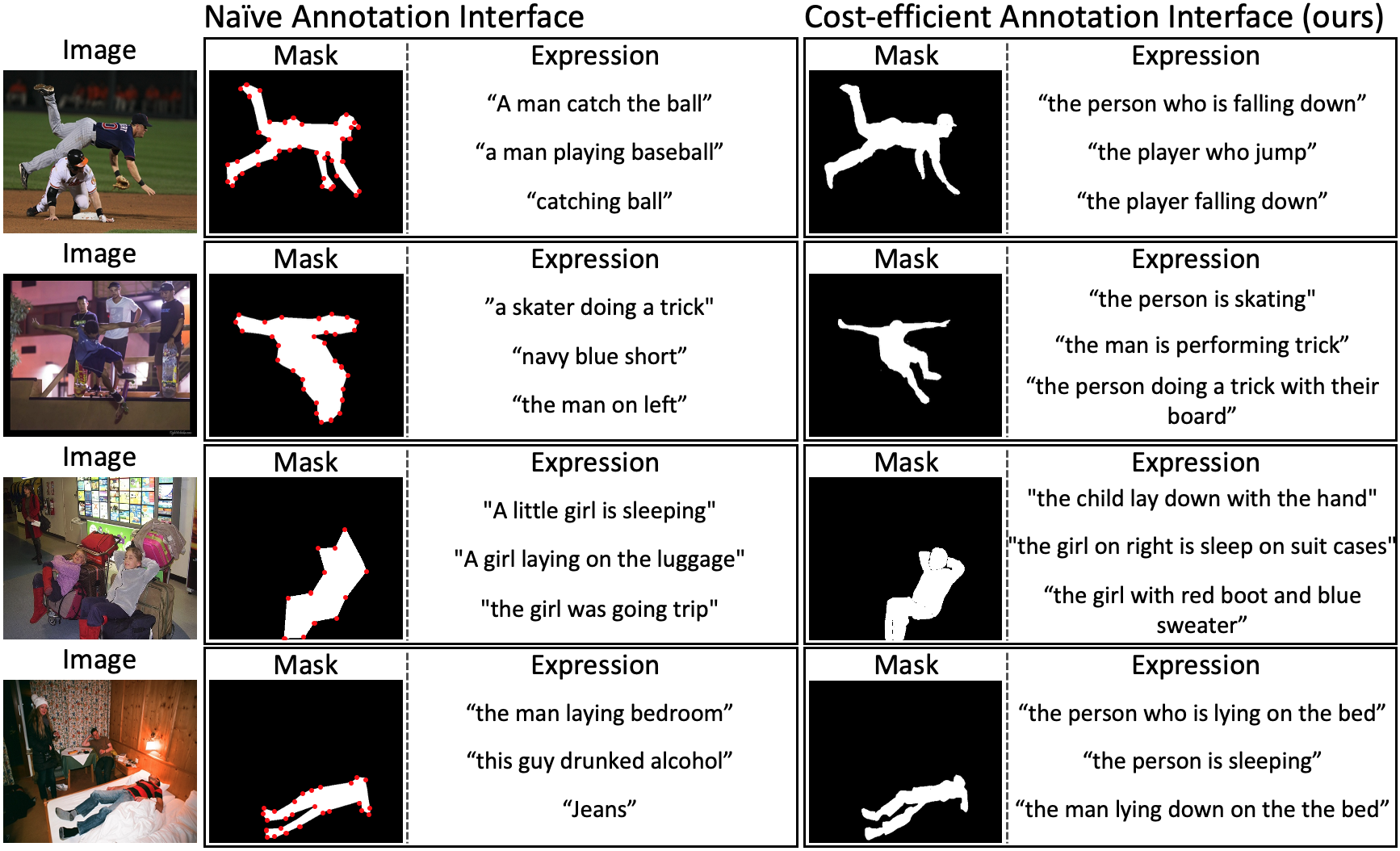}
    \caption{Examples of user study results with our cost-efficient annotation interface compared to the na\"ive annotation interface (\ie, polygon drawing for masks and manual typing for expressions).}
    \label{fig:supple_qual_user} 
\end{figure*}

To verify the efficiency of the proposed annotation interface, we conduct a user study with 30 annotators from Amazon Mechanical Turk in Tab.~\ref{tab:user_study} of the main manuscript.
In this subsection, we provide the details of the user study.
Each annotator is required to complete annotations for 10 images, following the instructions in Fig.~\ref{supfig:annotation_instructions}; the instructions also include example videos for easy understanding.
We record the time required to annotate both the mask and referring expression and save the annotation results (Fig.~\ref{fig:supple_qual_user}) to compare the quality with the ground truth labels.
Besides our proposed interface, we also conduct the same user study for a na\"ive annotation (\ie, polygon drawing for masks and manual typing for expressions) to evaluate the benefits of our proposed tool.
The example videos using our labeling interface are provided in our \href{https://github.com/junbum766/ALRIS}{GitHub repository}.

\subsection{Dataset}
\label{app:subsec:dataset}
\paragraph{RefCOCO.}
RefCOCO~\cite{refcoco} provides images paired with masks or boxes and their corresponding referring text.
This is created using the ReferItGame~\cite{referit}, where one player writes a textual description for a specific instance, and a second player locates the correct object based on that description.
The dataset contains 142,210 expressions referring to 50,000 instances across 19,994 images. 
These instances are divided into the following splits: 42,404 for a training set, 3,811 for a validation set, 1,975 for a testA set, and 1,810 for a testB set.
The expressions in RefCOCO are typically concise and often include directional instructions, such as ``\textit{left}'' or ``\textit{right}'', to disambiguate similar objects.

\paragraph{RefCOCO+.}
RefCOCO+~\cite{refcoco} is constructed using the same ReferItGame~\cite{referit} method as RefCOCO.
The difference in textual form is that it explicitly prohibits the use of positional terms. 
This constraint forces the expressions to be based primarily on object attributes, such as clothing color or object size.
This distinction allows researchers to evaluate a model's performance on attribute comprehension, separate from its understanding of spatial cues.
It includes 141,564 expressions for 49,856 instances across 19,992 images. 
These instances are divided into 42,278 for a train set, 3,805 for a val set, 1,975 for a testA set, and 1,798 for a testB set, following a similar split structure to RefCOCO.

\paragraph{RefCOCOg.}
RefCOCOg~\cite{refcoco_umd} is characterized by much longer expressions, averaging 8.43 words per expression, compared to 3.61 and 3.53 words for RefCOCO and RefCOCO+, respectively.
Therefore, it demands a model's ability for more complex and contextual comprehension and is commonly used to evaluate multi-modal reasoning under longer linguistic context.
Compared to RefCOCO and RefCOCO+, which are built on ReferItGame~\cite{referit}, RefCOCOg is collected via Amazon Mechanical Turk.
This dataset includes 95,010 expressions for 49,822 object instances across 25,799 images. These instances are separated into 42,226 for a train split, 2,573 for a validation split, and 5,023 for a test split.

\begin{figure*}[h!]
    \centering
    \includegraphics[width=0.98\linewidth]{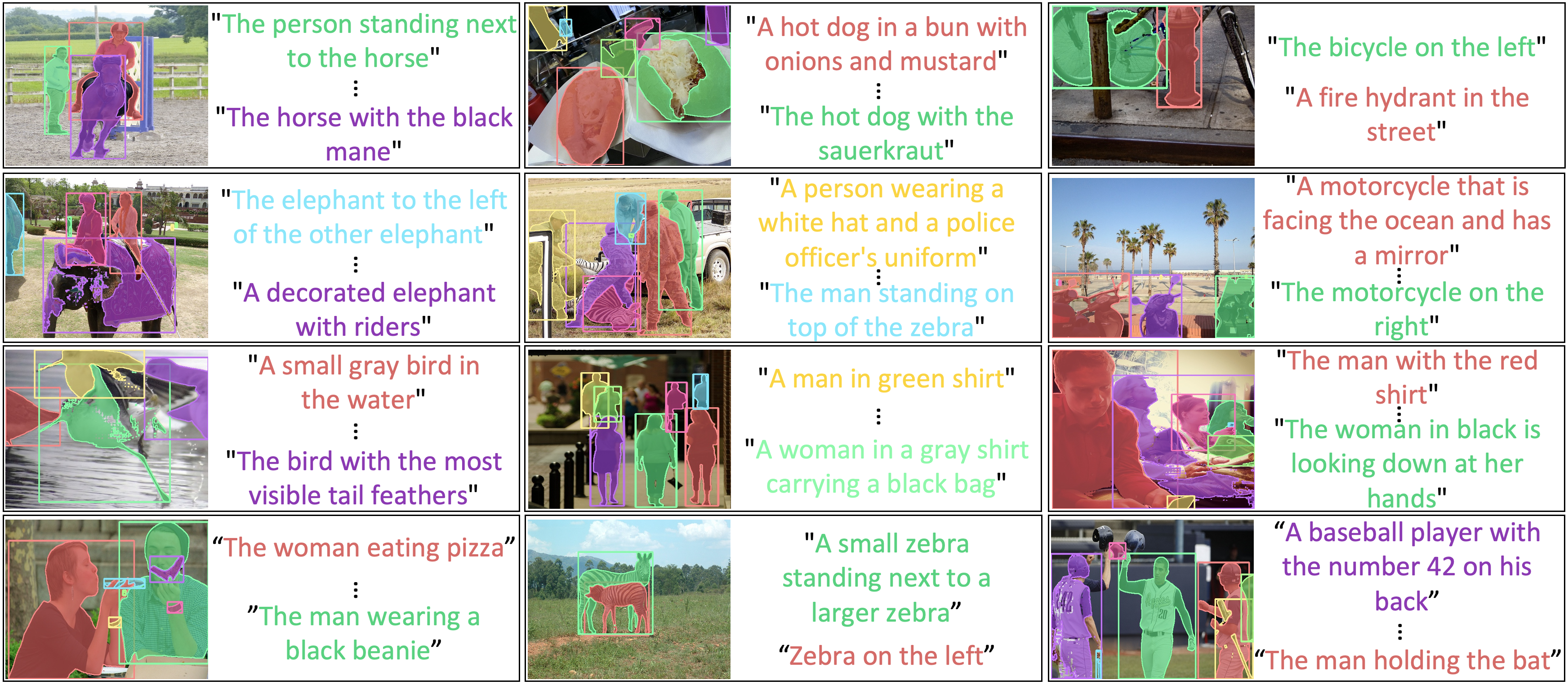}
    \caption{Examples of the generated auxiliary region-text pairs. The paired region-text are denoted by matching colors.}
    \label{fig:supple_qual_aux} 
\end{figure*}

\begin{figure}[t]
    \centering
    \includegraphics[width=1\linewidth]{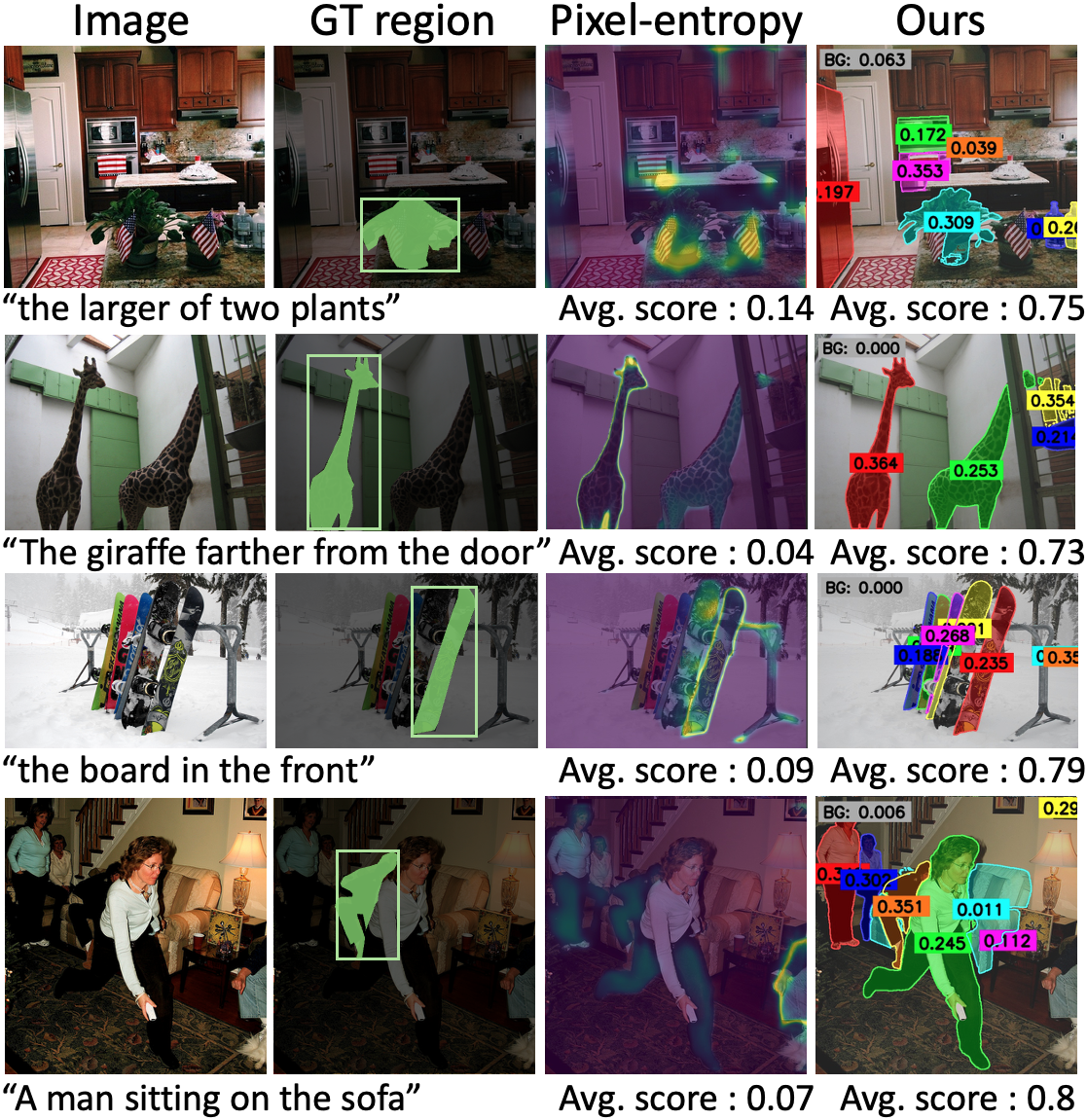}
    \caption{Additional qualitative analysis on the process of computing the informativeness score compared to the pixel-entropy.
    } 
    \label{fig:supple_qual_score} 
\end{figure}

\section{Additional Qualitative Results}
\label{supple:additional_qual}

\subsection{Examples of User Study Results}
\label{app:subsec:user_study}
We conduct a user study with 30 annotators from Amazon Mechanical Turk in Tab.~\ref{tab:user_study} of the main manuscript.
In Fig.~\ref{fig:supple_qual_user}, we provide qualitative examples of those results compared with the na\"ive labeling tool (\ie, polygon drawing for masks and manual typing for expressions).
The masks annotated via our interface are more accurate than those drawn manually, since polygon drawing is labor-intensive and struggles with complex object boundaries.
Our mask tool leverages precise pre-extracted SAM masks, enabling high-quality annotations with faster times.
Similarly, the expressions annotated through our interface accurately describe the referred regions.
This is confirmed by the higher Sentence-BERT similarity scores compared to manual typing in Tab.~\ref{tab:user_study} of the main manuscript.
This is because, at each step, the decoding process of the multi-modal language model (\ie, ViP-LLaVA~\cite{vip_llava}) suggests contextually appropriate words relevant to the target region, guiding annotators to concise and accurate descriptions.

\subsection{Examples of Auxiliary Region-Text Pairs}
\label{app:subsec:auxiliary}
We provide qualitative examples of the generated auxiliary region-text pairs (Sec.~\ref{subsec:pseudo_data} of the main manuscript) in Fig.~\ref{fig:supple_qual_aux}, where each region-text pair is highlighted using the same colors.
Although some pairs contain minor noise, the auxiliary regions accurately localize the objects, and the corresponding captions provide clear and detailed referring expressions for each region, as quantified in Tab.~\ref{tab:quality_auxiliary}.

\begin{figure*}[t]
    \centering
    \includegraphics[width=0.95\linewidth]{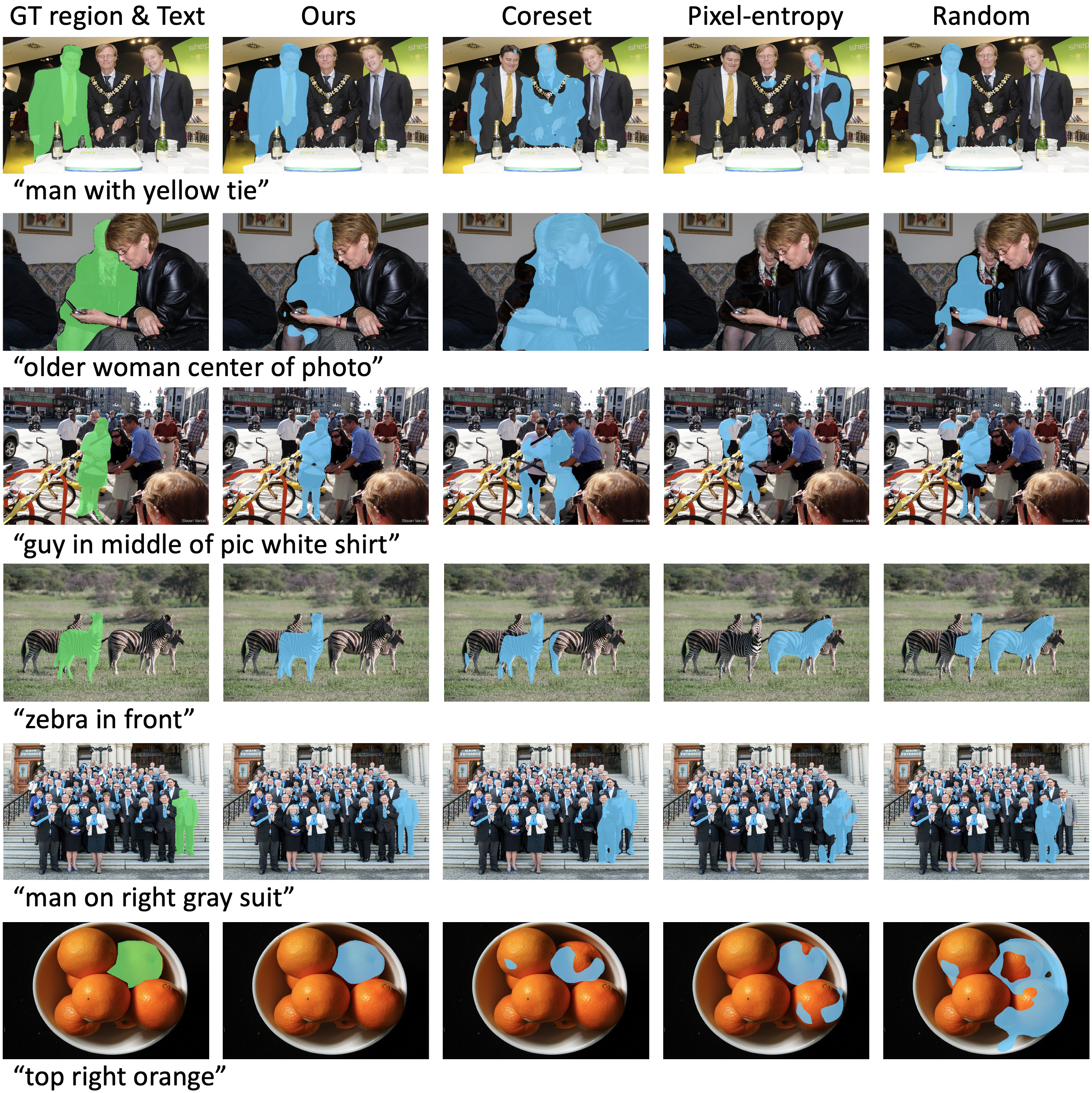}
    \caption{Qualitative examples from the ambiguous subset compared with the RIS models trained under different acquisition strategies.}
    \label{fig:supple_ambiguous} 
\end{figure*}

\subsection{Examples on Ambiguous Subset}
In Tab.~\ref{tab:ambiguous_subset} of Appendix~\ref{supple:ambiguous_subset}, we conduct comparisons with several acquisition baselines (\ie, random, coreset, pixel-entropy, and ours) on the curated ambiguous test subset.
In this subsection, we present their qualitative results from this ambiguous subset in Fig.~\ref{fig:supple_ambiguous}.
Compared with other acquisitions, the RIS model trained under our acquisition performs better at distinguishing the target from visually similar distractors.
This result suggests that the proposed referred region ambiguity is effective in improving the model's discriminative ability in challenging ambiguous cases.

\subsection{Examples of Selected Samples via Acquisitions}
\label{app:subsec:selected_samples}
\begin{figure*}[t] 
    \centering
    \hfill
    \begin{subfigure}{0.31\textwidth} 
        \centering
        \includegraphics[width=\linewidth]{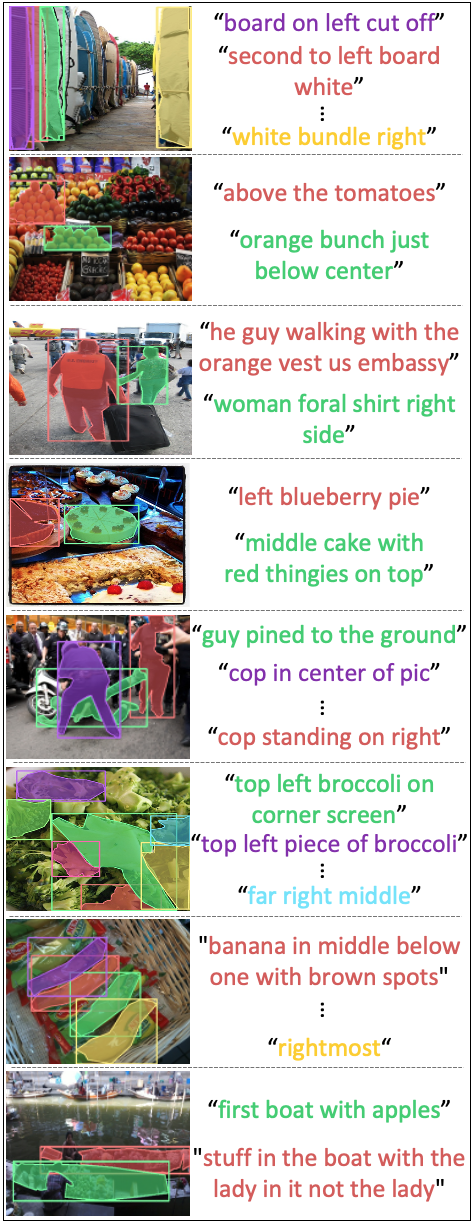}
        \caption{Referred region ambiguity}
        \label{fig:supple_acq_qual_a} 
    \end{subfigure}
    \hfill 
    \begin{subfigure}{0.31\textwidth}
        \centering
        \includegraphics[width=\linewidth]{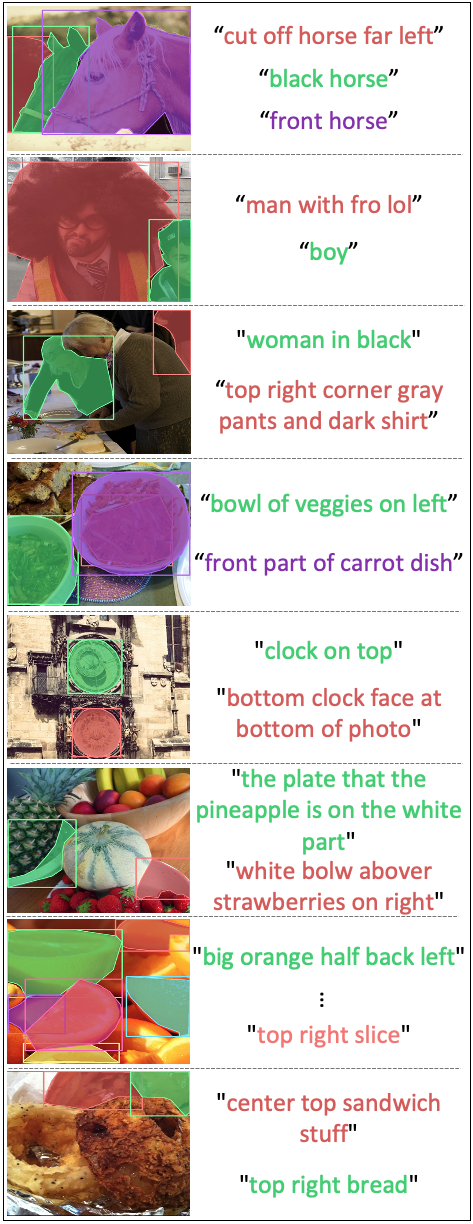}
        \caption{Pixel-entropy}
        \label{fig:supple_acq_qual_b} 
    \end{subfigure}
    \hfill
    \begin{subfigure}{0.31\textwidth}
        \centering
        \includegraphics[width=\linewidth]{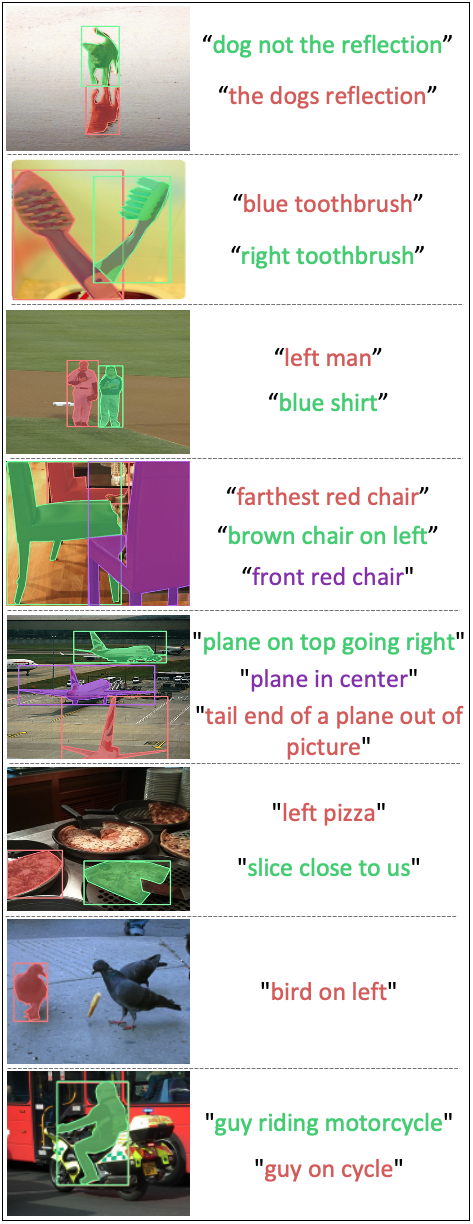}
        \caption{Coreset}
        \label{fig:supple_acq_qual_c} 
    \end{subfigure}
    \hfill
    \caption{Additional examples of selected samples via acquisitions. The paired GT region-text are denoted by matching colors.} 
    \label{fig:supple_acq_qual} 
\end{figure*}

In Fig.~\ref{fig:supple_acq_qual}, we present additional examples of selected samples via each acquisition.
Compared with other approaches (\ie, pixel-entropy and coreset), our referred region ambiguity tends to prioritize images containing complex scenes with many objects, leading to more informative selections.

\subsection{Examples of Referred Region Ambiguity}
\label{app:subsec:qual_score}
We show additional examples of the process for computing the informativeness score in our referred region ambiguity, compared with the pixel-entropy in Fig.~\ref{fig:supple_qual_score}.
Although the images contain visually similar objects, pixel-entropy assigns low informativeness because it averages entropy over all pixels. 
In contrast, our acquisition function yields high scores by operating at the region level, effectively capturing the complexity of these scenes.

\end{document}